\documentclass[letterpaper]{article}
\usepackage{aaai}
\usepackage{times}
\usepackage{helvet}
\usepackage{courier}
\usepackage{subfigure}
\usepackage{epsfig}
\usepackage{graphicx}
\begin{document}
%
\title{The Interpretable Dictionary in Sparse Coding}
\author{Edward Kim \\Computer Science \\ Drexel University\\
Philadelphia, PA 19104\\
ek826@drexel.edu\\
\And
Connor Onweller
\\Computer and Information Sciences \\ University of Delaware\\
Newark, DE 19716 \\
onweller@udel.edu\\
\And
Andrew O'Brien
\\Computer Science \\ Drexel University\\
Philadelphia, PA 19104\\
ao543@drexel.edu\\
\And
Kathleen McCoy
\\Computer and Information Sciences \\ University of Delaware\\
Newark, DE 19716 \\
mccoy@udel.edu\\}

\maketitle
\begin{abstract}
\begin{quote}
Artificial neural networks (ANNs), specifically deep learning networks, have often been labeled as black boxes due to the fact that the internal representation of the data is not easily interpretable.  In our work, we illustrate that an ANN, trained using sparse coding under specific sparsity constraints, yields a more interpretable model than the standard deep learning model.  The dictionary learned by sparse coding can be more easily understood and the activations of these elements creates a selective feature output.  We compare and contrast our sparse coding model with an equivalent feed forward convolutional autoencoder trained on the same data.  Our results show both qualitative and quantitative benefits in the interpretation of the learned sparse coding dictionary as well as the internal activation representations.  
\end{quote}
\end{abstract}

\section{Introduction}
Interpretability in machine learning has been defined as the degree to which a human can understand the cause of a decision \cite{miller2019explanation}.  In the case of a simple linear model with few input dimensions, the interpretability of this class of mathematical model is fairly transparent.  However, with the advent of deep learning models that can handle millions of dimensions, with millions of parameters in a highly non-linear, non-convex space, the interpretability of such models is virtually non-existant to the point that the community refers to them as black boxes.  

Rudin argues that attempting to explain black box models is problematic \cite{rudin2019stop}; explanations on top of an uninterpretable model are not reliable, and could be misleading.  This is because creating a different explainable module is generally accomplished by running a simpler (explainable) model concurrently to the existing model. Such a model cannot have perfect fidelity with respect to the original model.  If it could, it would be equal to the original model, and one would not need the original model in the first place.  Therefore, any type of separate model that attempts to explain the actions of another model is inherently flawed.   

If we cannot explain a deep learning black box model, should we just give up on these types of models trained via backpropagation and start over?  In a 2017 interview, Geoffery Hinton, one of the founding pioneers of deep learning, suggested exactly this.  ``I don't think it's how the brain works... My view is throw it all away and start again,'' says Hinton.   In our work, we share the view that the human brain gives us insight on designing an improved learning system.  We also believe that the only way to create an interpretable model is to start with interpretable components.  We need to design models that are composed of inherently interpretable elements, which would then be interpretable as a whole \cite{rudin2019stop}.  

In the following work, we present a method for training a neural network so that the components of the network are interpretable.  Our work is inspired by representation and learning mechanisms that we observe in the mammalian brain.  Several concepts employed by the brain, such as attention, competition, and sparsity, naturally emerge from intelligence in part because they help to explain the world around us.  In our experiments and results, we illustrate how our framework compares to an equivalent feed-forward deep learning model that would commonly be used in the state-of-the-art machine learning models today.
\section{Background}

\subsection{Background in Attention in the Brain}
Neuroscientists have theorized that the brain needs to be selective about what information it chooses to process due to the overwhelming amount of available information \cite{knudsen2007fundamental}. One selection mechanism used by the brain is to bias the response of neurons to certain stimuli. In a landmark study, Hubel and Wiesel demonstrated that a cat’s neural firing was related to the placement, orientation, and direction of movement of a bright line being moved across its retina \cite{hubel1962receptive}. Oriented edges, spatial frequency, color contrast, intensity, and motion have been identified as key determinants of whether visual stimuli attract attention \cite{itti2001feature}.

These neurons in the cortex are exhibiting selectivity to different stimuli.  We see similar patterns of stimulus selectivity at higher levels of the brain i.e. specifically in the inferior temporal gyrus or IT, where regions are selective to specific objects, faces, body parts, places, and words.  At the level of the cortex, this means that a high majority of neurons are not being activated simultaneously.  In fact, many neurons are silent most of the time, with a single percent to a few percent of neurons spiking at any one time.  Given a specific stimulus, only a highly selective, small subset of neurons will activate \cite{shoham2006silent}.

\subsection{Background in Convolutional Neural Networks}
In machine learning, the convolutional neural network was originally inspired by studying biological visual processing.
The motivation behind these networks was to develop
a reliable, self-taught pattern recognition system by modeling the neurological
structures related to pattern recognition found in vertebrates. The
``neocognitron'' model presented by Kunihiko Fukushima in 1980, which is
regarded as the earliest convolutional neural network, took great inspiration
from the work of neurophysiologists Hubel and Wiesel. In fact, Fukushima modeled
his network after the hierarchical model of the visual cortex in which cells later in the hierarchy have increasingly large
receptive fields and respond to more complex stimulus patterns \cite{fukushima1980}.

LeCun modified this architecture with the creation of LeNet \cite{lecun1989}.  This model applied supervised learning techniques to train a similar 
hierarchical network structure for optical character recognition.
Instead of having the network assign activations
to its units via repeated exposure to stimulus patterns, they used
backpropagation to train the network to classify characters using labeled
examples. A typical CNN is a hierarchical network made up of
several convolutional layers and pooling layers and is trained via some supervised
learning technique \cite{lecun2015}.  Nearly all CNNs today follow this archetype, i.e. supervised learning with backpropagation.

\subsection{Background in Sparse Coding}
Parallel to the CNN effort, some computational methods emerged that more faithfully model biological neural networks.  
Sparse coding was first introduced by Olshausen and Field \cite{olshausen1997sparse}, in order to explain how the primary visual cortex efficiently encodes natural images.  The key element of the neural code is that it reduces the redundancy encoded, such that the underlying signal contains independent entities useful for the survival of the organism.  In other words, given an overcomplete basis where the basis functions are non-orthogonal and not linearly independent, one must attempt to represent the input signal with a sparse vector where only a small number of basis functions are chosen out of a larger set \cite{field1994goal}.   

Mathematically, sparse coding can be formulated as a reconstruction minimization problem defined as follows.  In the sparse coding model, we have some input variable $x^{(n)}$ from which we are attempting to find a latent representation $a^{(n)}$ (we refer to as ``activations'') such that $a^{(n)}$ is sparse, e.g. contains many zeros, and we can reconstruct the original input, $x^{(n)}$ with high fidelity.  A single layer of sparse coding can be defined as,
\begin{equation}
	\min_\Phi \sum^{\mathcal{N}}_{n=1} \min_{a^{(n)}} \frac{1}{2} \| x^{(n)} - \Phi a^{(n)}\|^2_2 + \lambda \|a^{(n)}\|_1
	\label{eq:sparsecode}
\end{equation}
Where $\Phi$ is the overcomplete dictionary, and $\Phi a^{(n)} = \hat{x}^{(n)}$, or the reconstruction of $x^{(n)}$.  The $\lambda$ term controls the sparsity penalty, balancing the reconstruction versus sparsity term.  $\mathcal{N}$ is the total training set, where $n$ is one element of training.   $\Phi$ represents a dictionary composed of small kernels that share features across the input signal.

\subsection{Relationship between Convolutional Neural Networks and Convolutional Sparse Coding}

\begin{figure}[ht]
 \centering
    \subfigure[Illustration of how a feed forward neural network would generate the activation vector.]{\includegraphics[width=7cm]{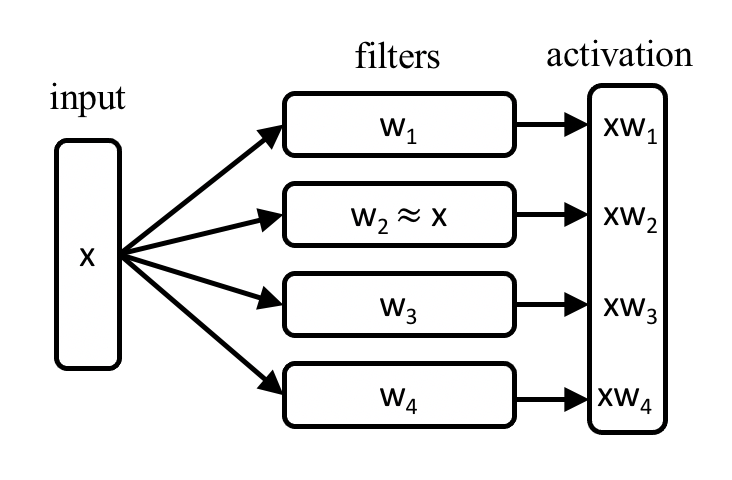}}\\
    \subfigure[Illustration of how a sparse coding algorithm would use lateral competition to generate the activation vector.]{\includegraphics[width=7cm]{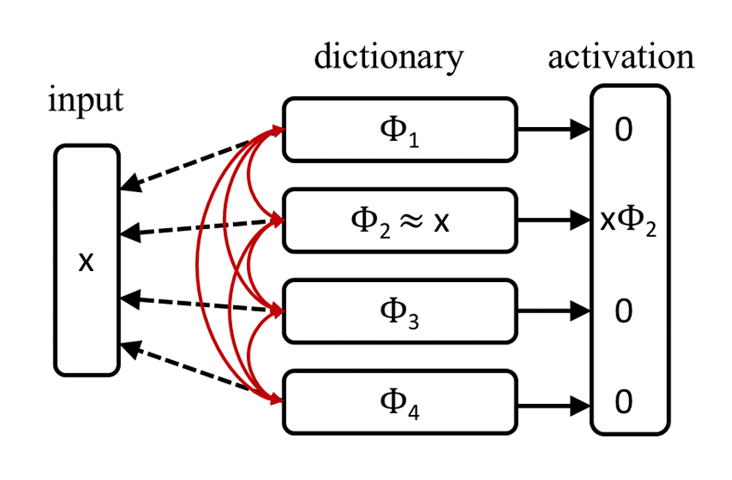}}
  \caption{An illustration of the mathematical process of generating an activation vector in a vectorized (a) feed forward convolutional neural network and a vectorized (b) sparse coding neural network.  Note that the second element ($w_2$ and $\Phi_2$), by chance, are approximately equal to the input vector, $x$.  In the feed forward case, all computations are independent from each other, while in the sparse coding case, every dictionary element needs to compete with one another to represent the input vector.  If one dictionary element ``explains'' the input sufficiently, all other activations will be zero.}

\label{fig:inverse}
\end{figure}

Given the history of deep learning, convolutional neural networks, and sparse coding, there are many similarities between the two methods.  However, the main difference between sparse coding and CNNs is that the CNN tries to solve the problem of feature extraction in a feed forward manner, whereas, sparse coding solves the problem in an inverse manner.  This is illustrated in Figure \ref{fig:inverse} and described mathematically below.

In the feed forward CNN, we can vectorize the input and weights of a convolutional neural network such that the primary computation can be expressed as a matrix vector multiplication,
\begin{equation}
    a = Wx 
\end{equation}
Where $a$ are the feature activations of the image filter vectorized as $W$ multiplied with the vectorized input, $x$.

In the case of sparse coding, we similarly vectorize the input and weights so that we solve the inverse problem.  
\begin{equation}
    x = \Phi a 
\end{equation}
Where $\Phi$ is the sparse coding dictionary and $\Phi^{-1}$ would be the equivalent of $W$ in the feed forward case needed to solve for the activations, $a$.   Some interesting hybrid architectures exist that attempt to bridge the gap between the two methods by
enforcing sparsity in a deep feed forward CNN \cite{quocle2012}. Unfortunately, sparse activations that employ lateral competition cannot be solved trivially, and requires an optimization under the assumption of sparsity.

\subsection{Interpretability Example} 
Imagine you walk into your living room and you see a broken window.  
You also see that all 10,000 of your children are running around in this room simultaneously.  You stop and yell, ``which one of you is responsible for this broken window?''  In the first scenario, all 10,000 of your children raise their hand, some more than others, but nonetheless, they all raise their hand.  Your spouse then enters the room and asks you, can you explain who is responsible?  We argue that the children's responses are neither interpretable nor explainable by an observer.  We contrast this with the second scenario.  You yell, `which one of you is responsible for this broken window?''  All 10,000 of your children begin to raise their hands; however, one child, Connor, says to everyone else, ``brothers and sisters, put your hands down, it was me.''  Everyone puts their hands down, and only Connor remains raising his hand.  In this situation, we are able to attend to a specific person, and attribute blame. Likewise, we argue in the remainder of this work that algorithms that employ competition between neural elements, and have activation maps that exhibit high sparsity and minimal cross correlation to other activation maps create interpretable units for a machine learning model. 

\section{Methodology}
In our research, we are interested in interpreting and explaining the content of an information graphic for use in assistive technology.  These information graphics, such as line graphs and bar graphs, are presented to the reader to support an intended message; however for someone who is visually impaired or otherwise unable to view the graphic, they will miss the information if it is not summarized in textual form.   Our previous work in information graphic classification describes a method for extracting the trends in a line graph \cite{kim2018multimodal} and creating textual summaries using a collection of deep learning models \cite{kim2020multimodal}.  However, given the limitations we faced with supervised deep learning models, we are now researching more interpretable models.  This image dataset consisting of 1,000 information graphics lends itself particularly well to the idea of interpretation because every element of an information graphic is purposefully and intentionally added by the author to the graphic for an explanatory reason.  

\subsection{Sparse Optimization}
 \begin{figure}[t]
    \subfigure[Initialization]{\includegraphics[width=2.75cm]{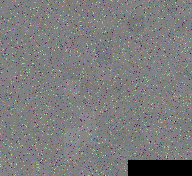}}
    \subfigure[2.5  Epochs]{\includegraphics[width=2.75cm]{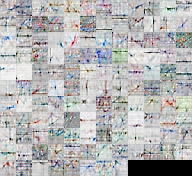}}
    \subfigure[20 Epochs]{\includegraphics[width=2.75cm]{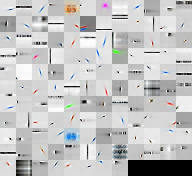}}
    
  \caption{Training under sparse optimization: (a) shows the initialization of $\Phi$ on the information graphics dataset.  (b) is the dictionary after 2.5 epochs of training.  (c) shows the fully converged dictionary, $\Phi$, after 20 epochs of training.}

\label{fig:dictionary}
\end{figure}
We use the Locally Competitive Algorithm (LCA) \cite{rozell2007locally} to minimize the mean-squared error in Equation \ref{eq:sparsecode}.  The LCA algorithm is a biologically informed sparse solver governed by dynamics that evolve the neuron's membrane potential when presented with some input stimulus.  Activations of neurons in this model laterally inhibit units within the layer to prevent them from firing.  The input potential to the state is proportional to how well the image matches the neuron's dictionary element, while the inhibitory strength is proportional to the activation and the similarity of the current neuron and competing neuron's convolutional patches, forcing the neurons to be decorrelated.  The LCA model is an energy based model similar to a Hopfield network \cite{hopfield1984neurons} where the neural dynamics can be represented by a nonlinear ordinary differential equation.  We define the internal state of a particular neuron as $u$ and the active coefficients as $a = T_{\lambda}(u)$, where $T$ is an activation function with threshold parameter, $\lambda$.  The dynamics of each node is determined by the set of coupled ordinary differential equations,
\begin{equation}
\label{eq:origdrive}
 \frac{du}{dt} =  -u + (\Phi^T x) - (\Phi^T \Phi a - a)
\end{equation}
where the equation is related to leaky integrators.  The $-u$ term is leaking the internal state of the neuron, the $(\Phi^T x)$ term is ``charging up'' the state by the inner product (match) between the dictionary elements and input patch, and the $(\Phi^T \Phi a - a)$ term represents the inhibition signal from the set of active neurons proportional to the inner product between dictionary elements.  The $- a$ in this case is eliminating self interactions.  In summary, neurons that match the input image charge up faster, then pass a threshold of activation.  Once they pass the threshold, other neurons in that layer are suppressed proportional to how similar the dictionary elements are between competing neurons.  This prevents the same image component from being redundantly represented by multiple nodes.  The dictionary elements can be learned by taking the gradient of the cost function with respect to $\Phi$, which leads to a local Hebbian learning rule that reduces reconstruction error given a sparse representation.  The dictionary elements are sized at 16x16 pixels, and the input images are resized to 128x128x3.  We further define 128 dictionary elements that stride over the input image by 4 pixels.  Thus the activation output is a 32x32x128 activation map.  We visualize the dictionary as it trains in Figure \ref{fig:dictionary}.

  \begin{figure}[t]
    \subfigure{\includegraphics[width=2.75cm]{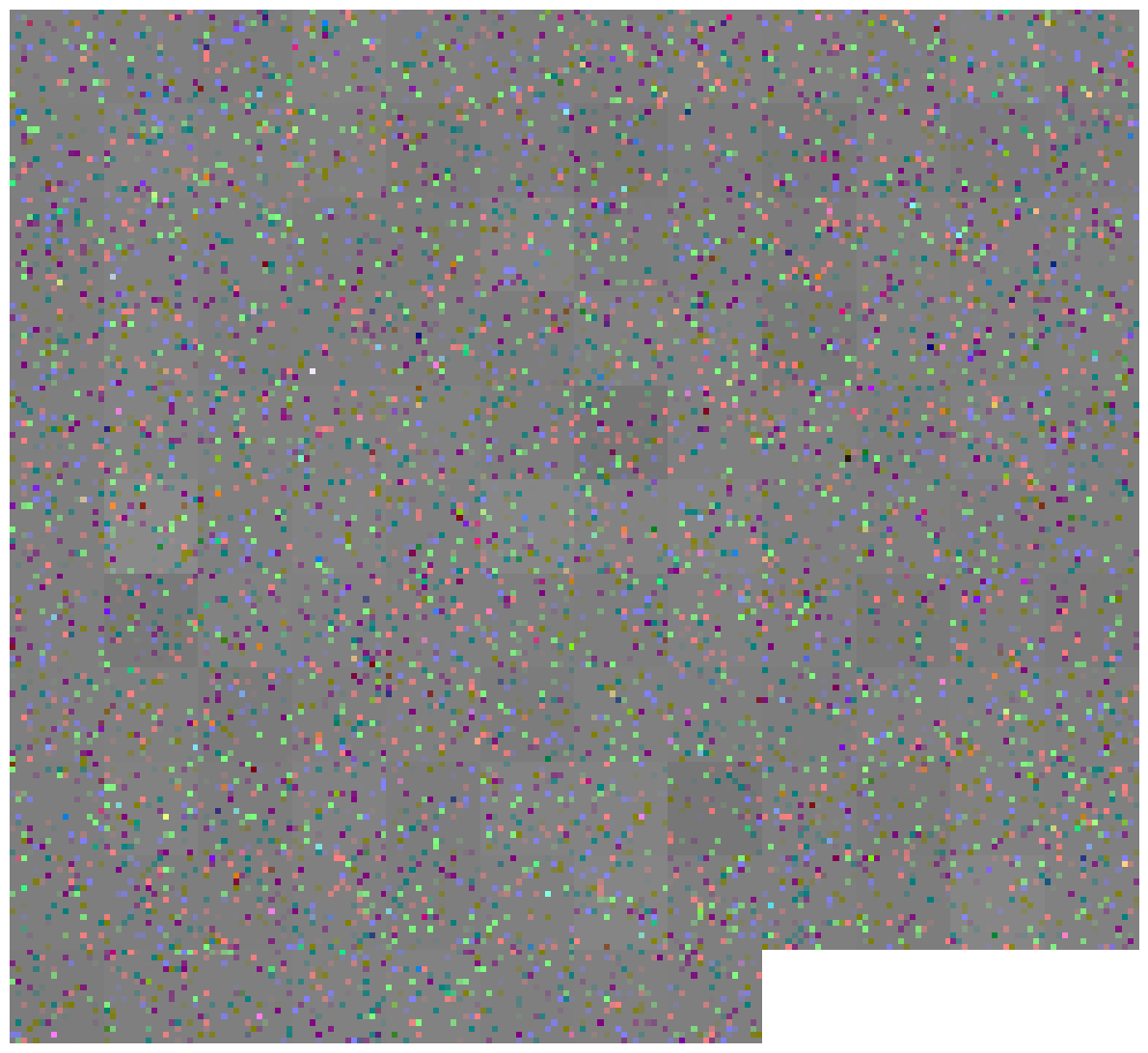}}
    \subfigure{\includegraphics[width=2.75cm]{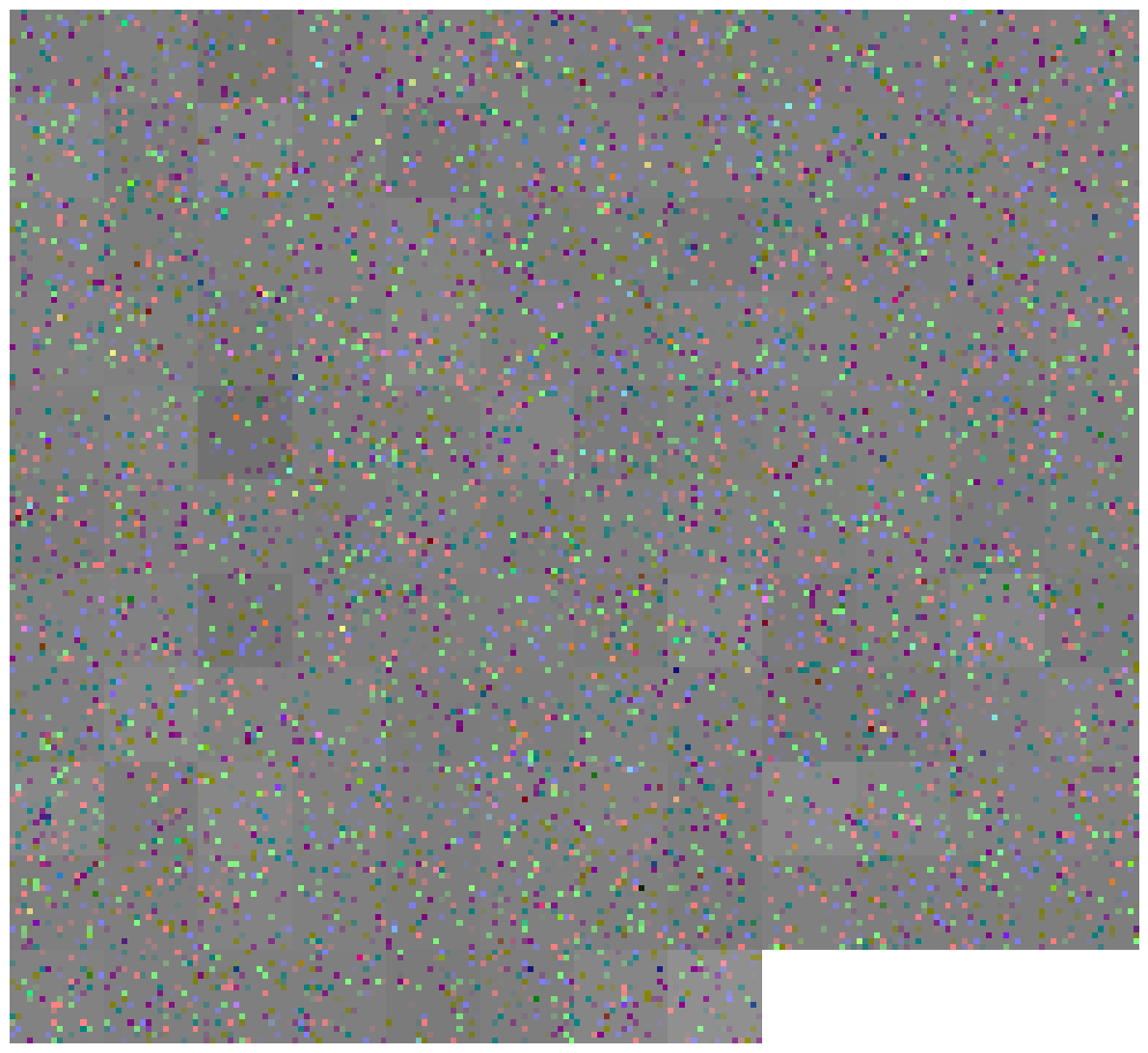}}
    \subfigure{\includegraphics[width=2.75cm]{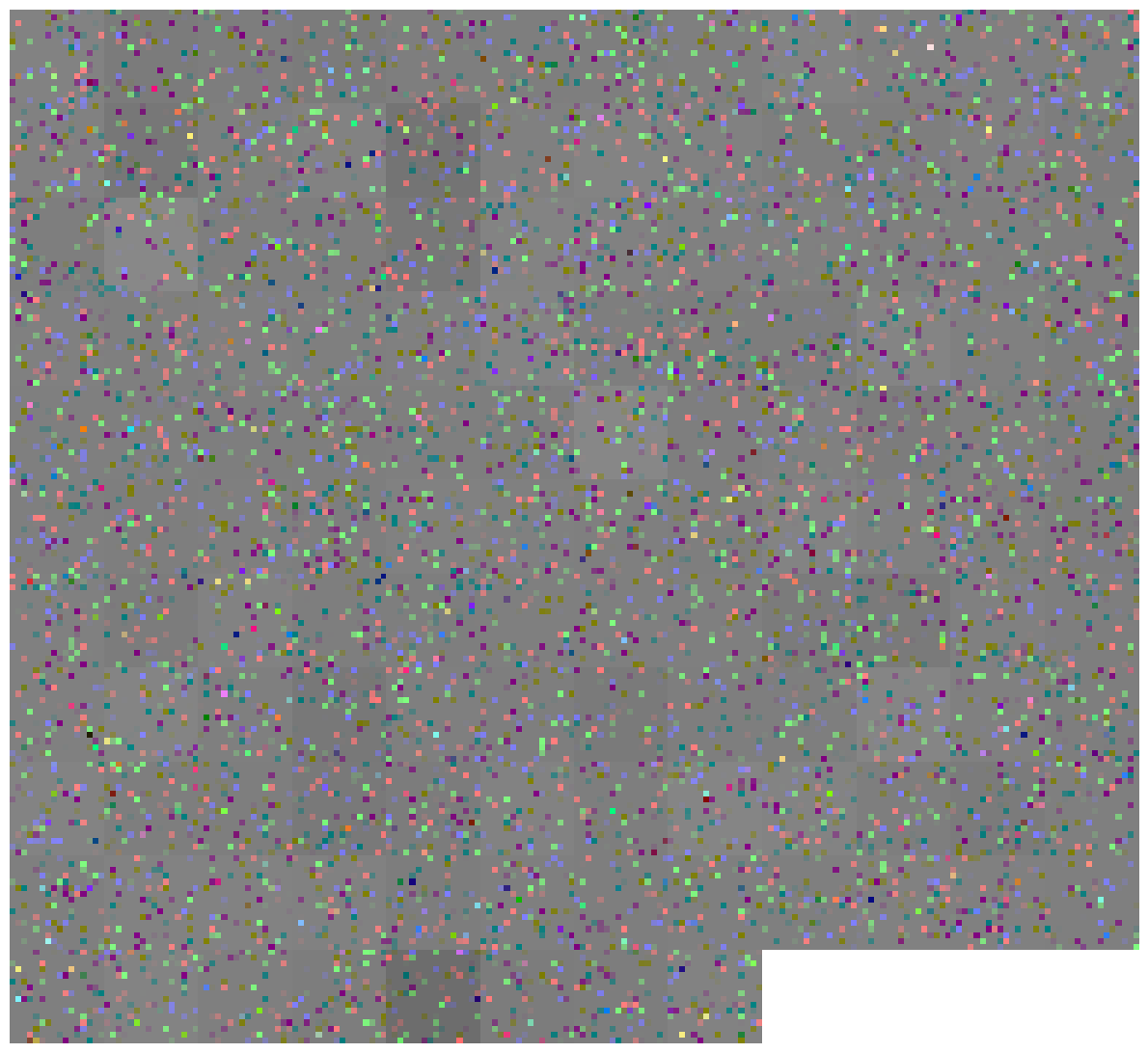}}
    \subfigure{\includegraphics[width=2.75cm]{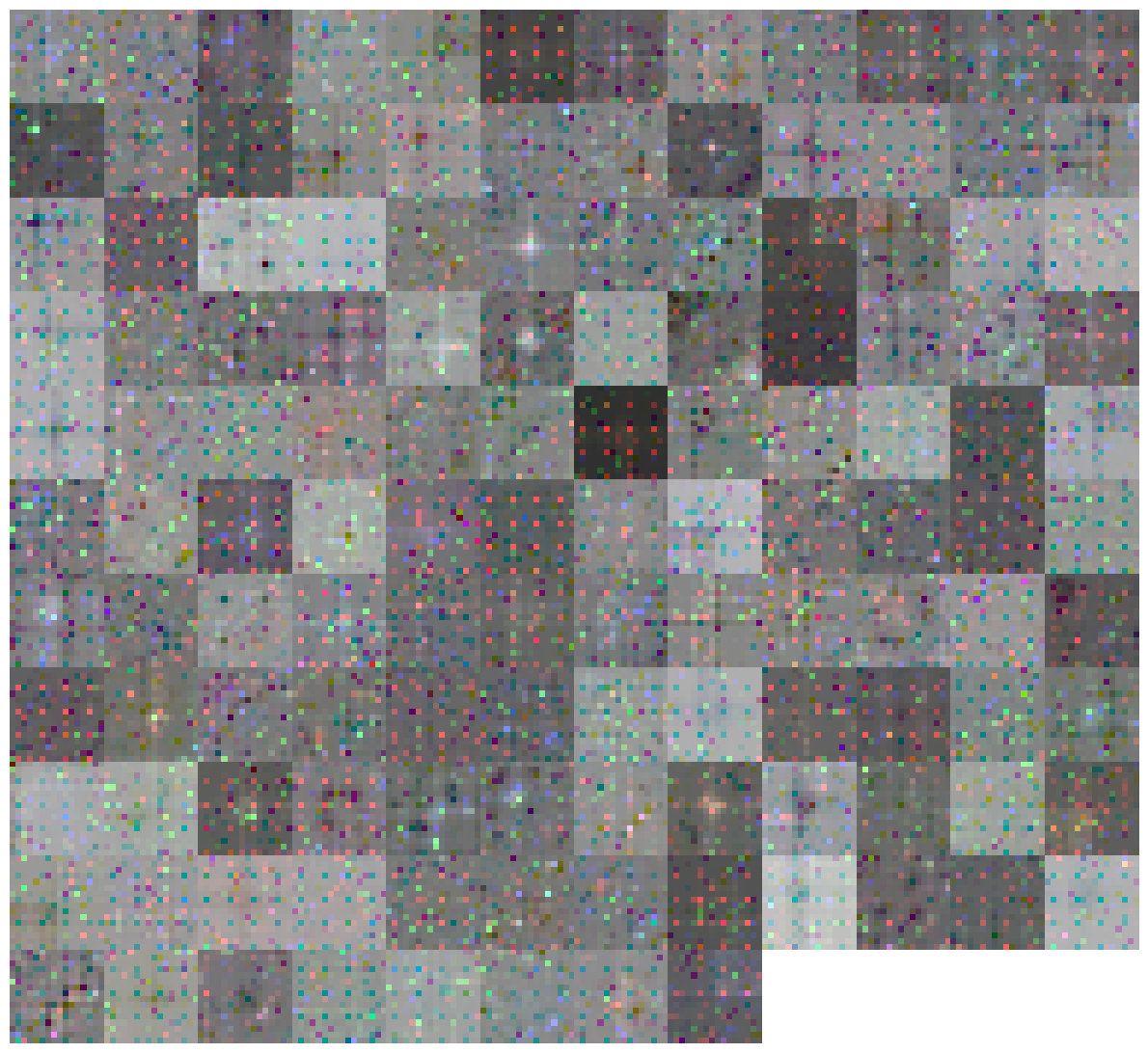}}
    \subfigure{\includegraphics[width=2.75cm]{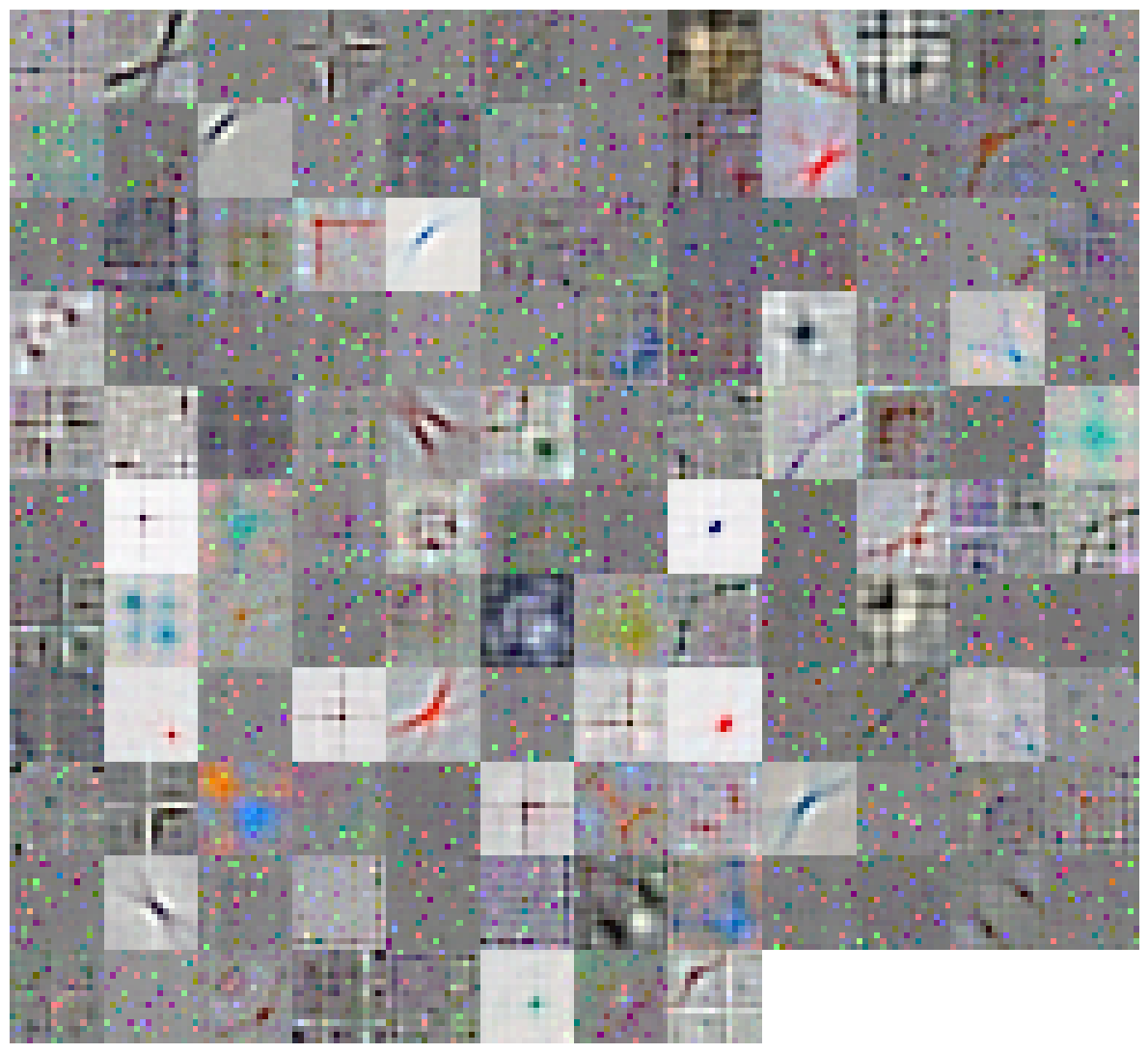}}
    \subfigure{\includegraphics[width=2.75cm]{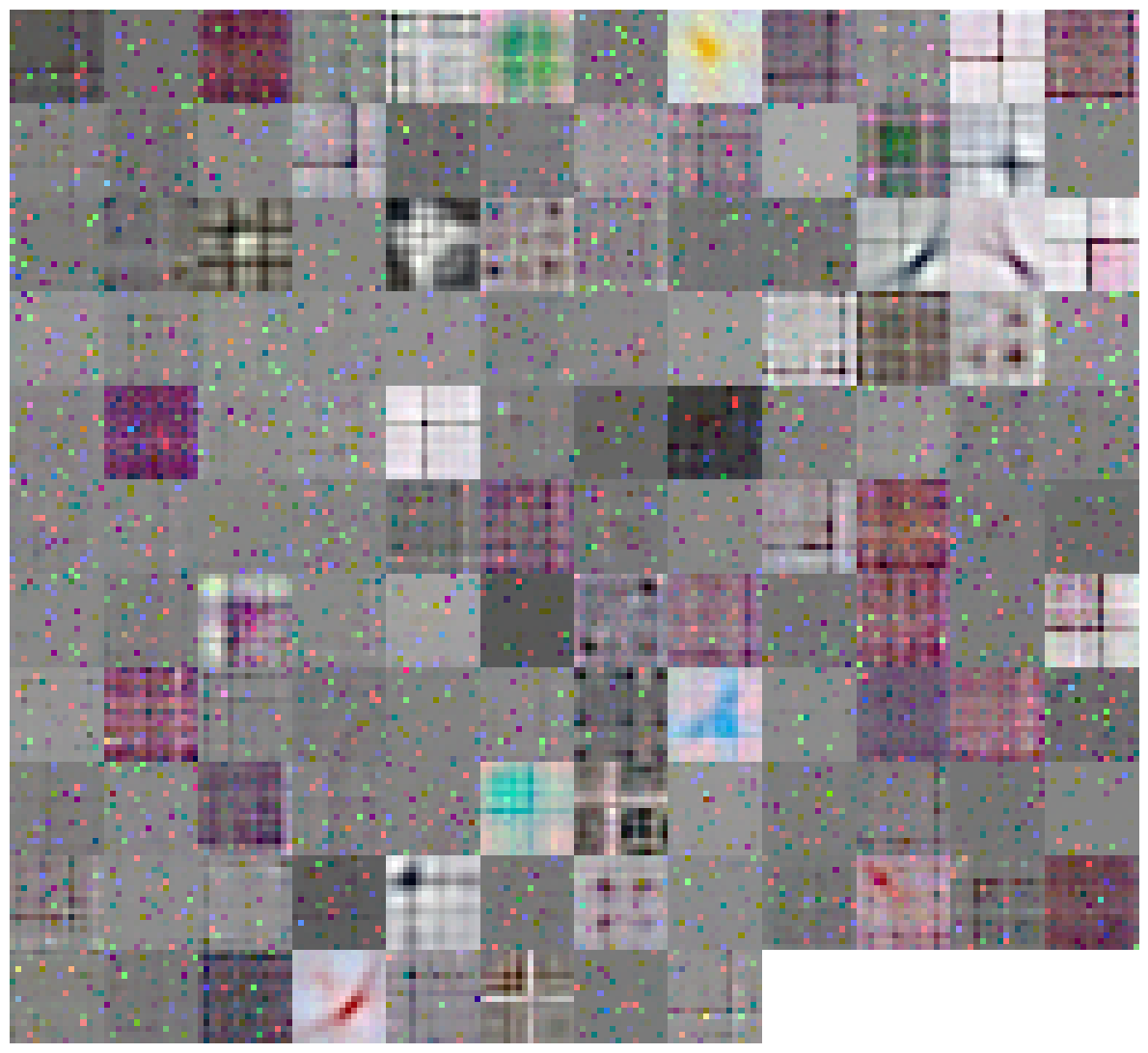}}
    \setcounter{subfigure}{0}\subfigure[No noise]{\includegraphics[width=2.75cm]{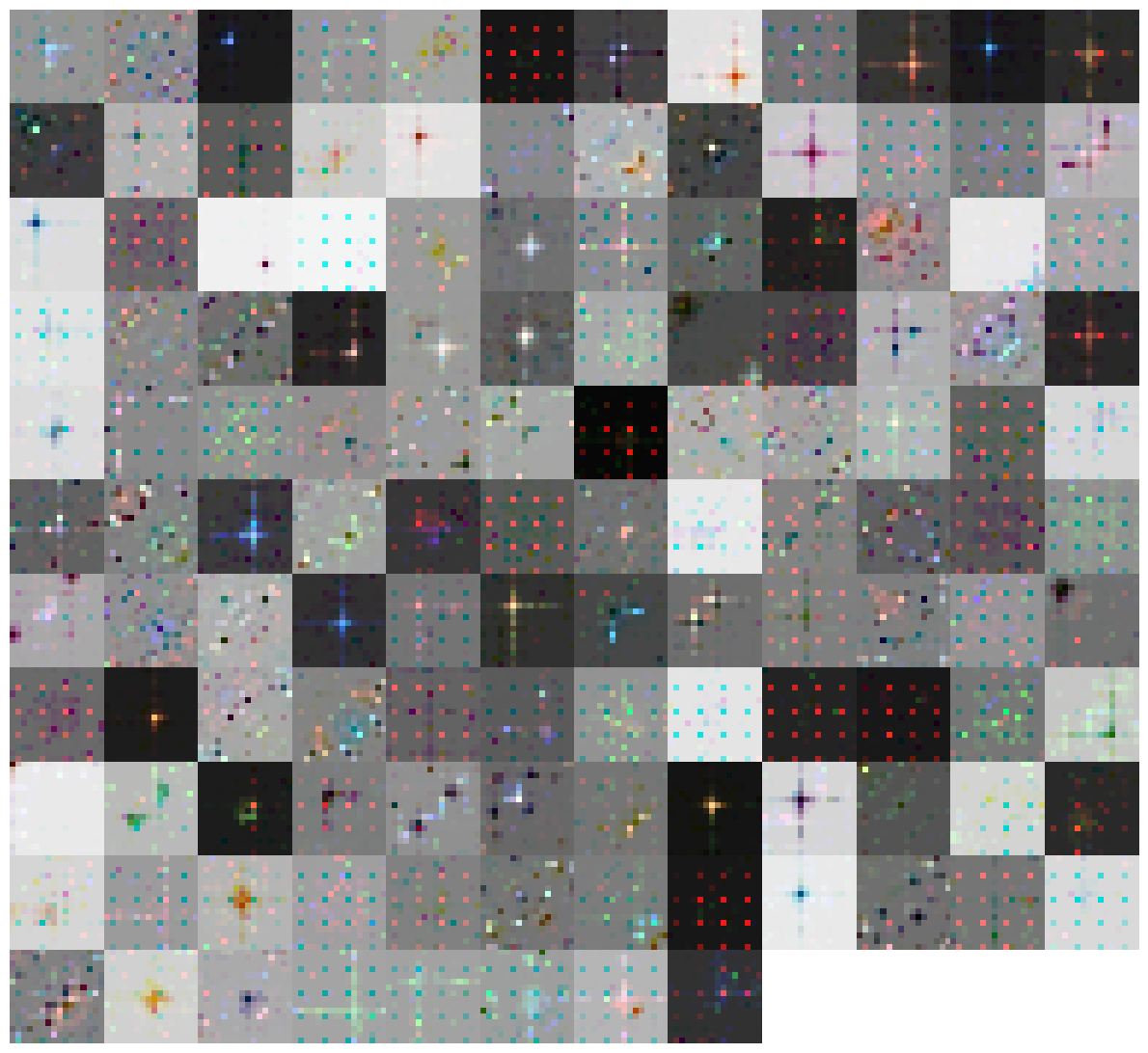}}
    \subfigure[Noise $\sigma$ = 0.5]{\includegraphics[width=2.75cm]{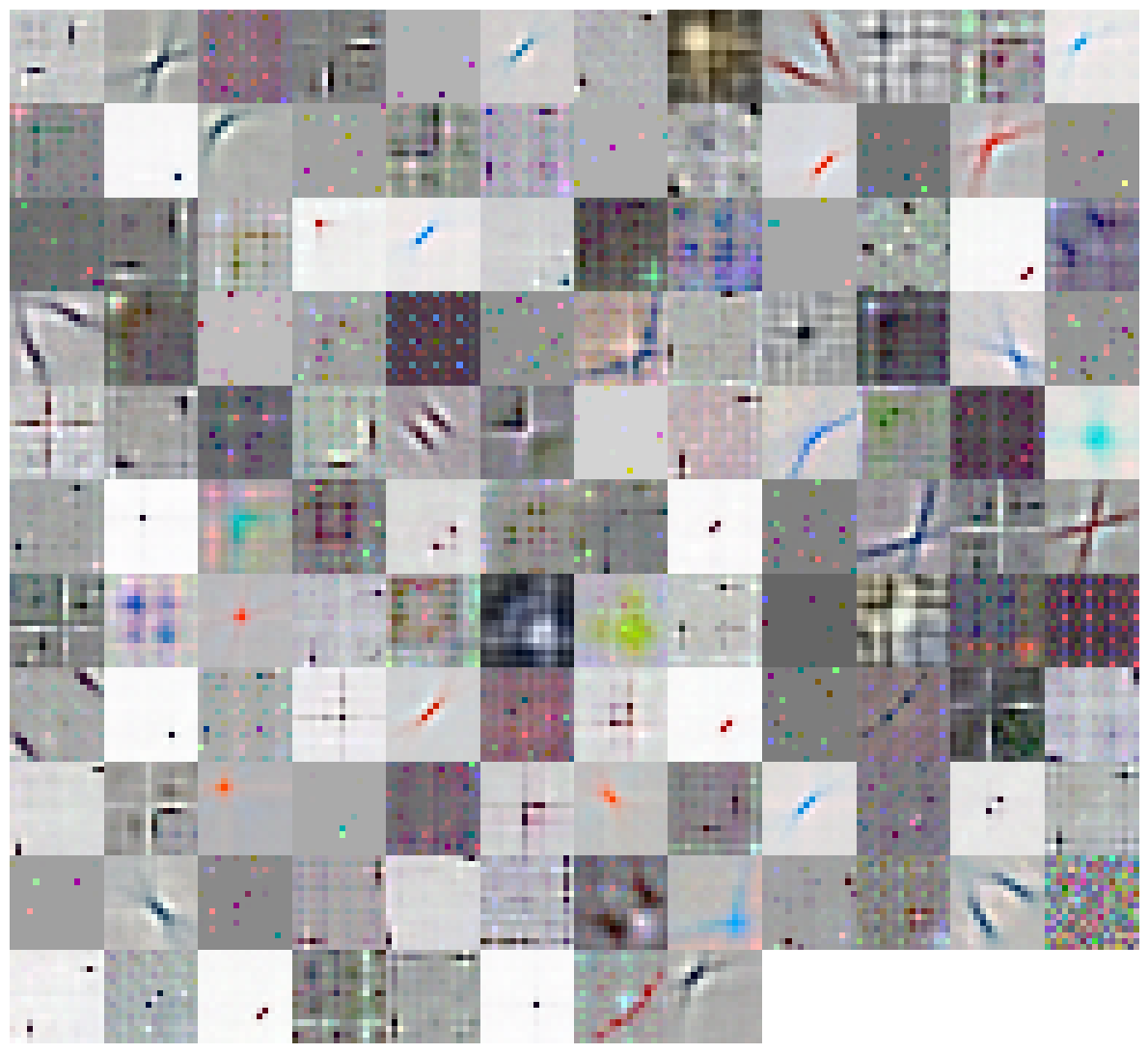}}
    \subfigure[Noise $\sigma$ = 1.0]{\includegraphics[width=2.75cm]{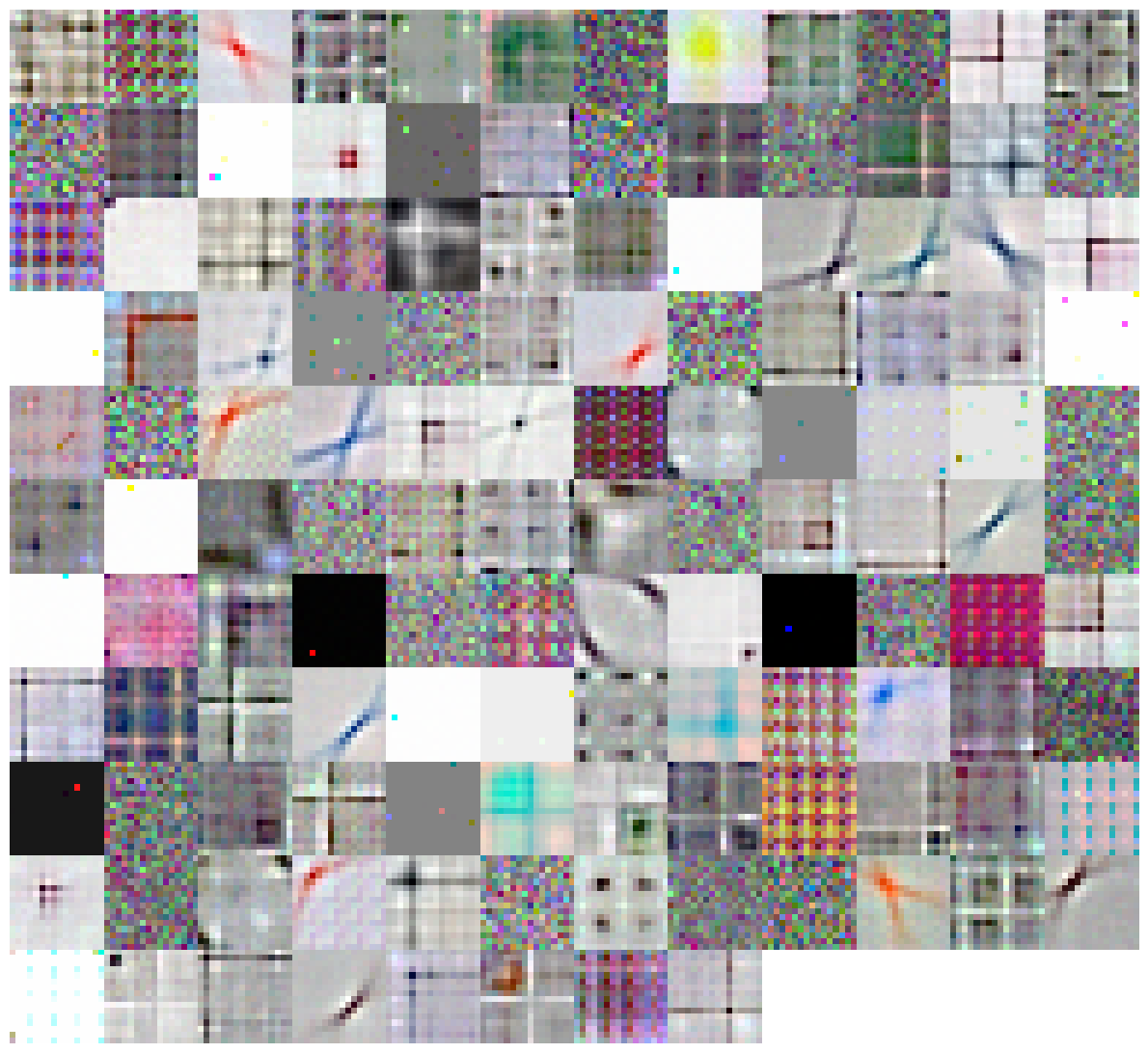}}
  \caption{CNN filters learned by an equivalent autoencoder architecture.  The first row is the initialization of the filters, the second row is the filters learned after 60 epochs through the data, the last row is the filters after 175 epochs.  The first column, (a), shows the features learned with no noise, the second column, (b), shows the features learned with $\sigma = 0.5$, and (c) shows several unstable features learned with $\sigma = 1.0$.}

\label{fig:filters}
\end{figure}

 \subsection{CNN Autoencoder}
 We implement a CNN autoencoder that is an equivalent architecture to our sparse coding formulation with a similar objective function.  We define 128 filters sized at 16x16 that stride over the input image by 4 pixels.  The activation maps of the CNN are of equivalent size as sparse coding, e.g. 32x32x128.  However, given the equivalency, we describe that a standard CNN autoencoder required some modification.  
 
 As noted by \cite{vincent2010stacked}, 
a standard autoencoder does not yield any significant structure.  Vincent et al. notes that both a regular over-complete autoencoder (200 hidden units) and an under-complete autoencoder
learn uninteresting features.  The under-complete case learns local blob detectors while the overcomplete case does not learn anything recognizable.  Our experiments yield the same phenomenon where the filters in our over-complete autoencoder do not learn semantic features as seen in Figure \ref{fig:filters}(a).  
 
We add a denoising criteria to the autoecoder to guide the learning to useful representations, similar to \cite{vincent2010stacked}.  Adding Gaussian white noise with $\sigma = 1.0$, semantic features did emerge; however, the learning became unstable after a high number of epochs (approximately 120 epochs), see Figure \ref{fig:filters}(c).  Our best stable features were learned with a denoising autoencoder with additive Gaussian white noise of $\sigma = 0.5$.  We observe several Gabor-like edge detectors as well as line segments that correspond to the information graphic dataset, see Figure \ref{fig:filters}(b).

\section{Results}

  \begin{figure}[th]
  \centering
    \includegraphics[width=6.5cm]{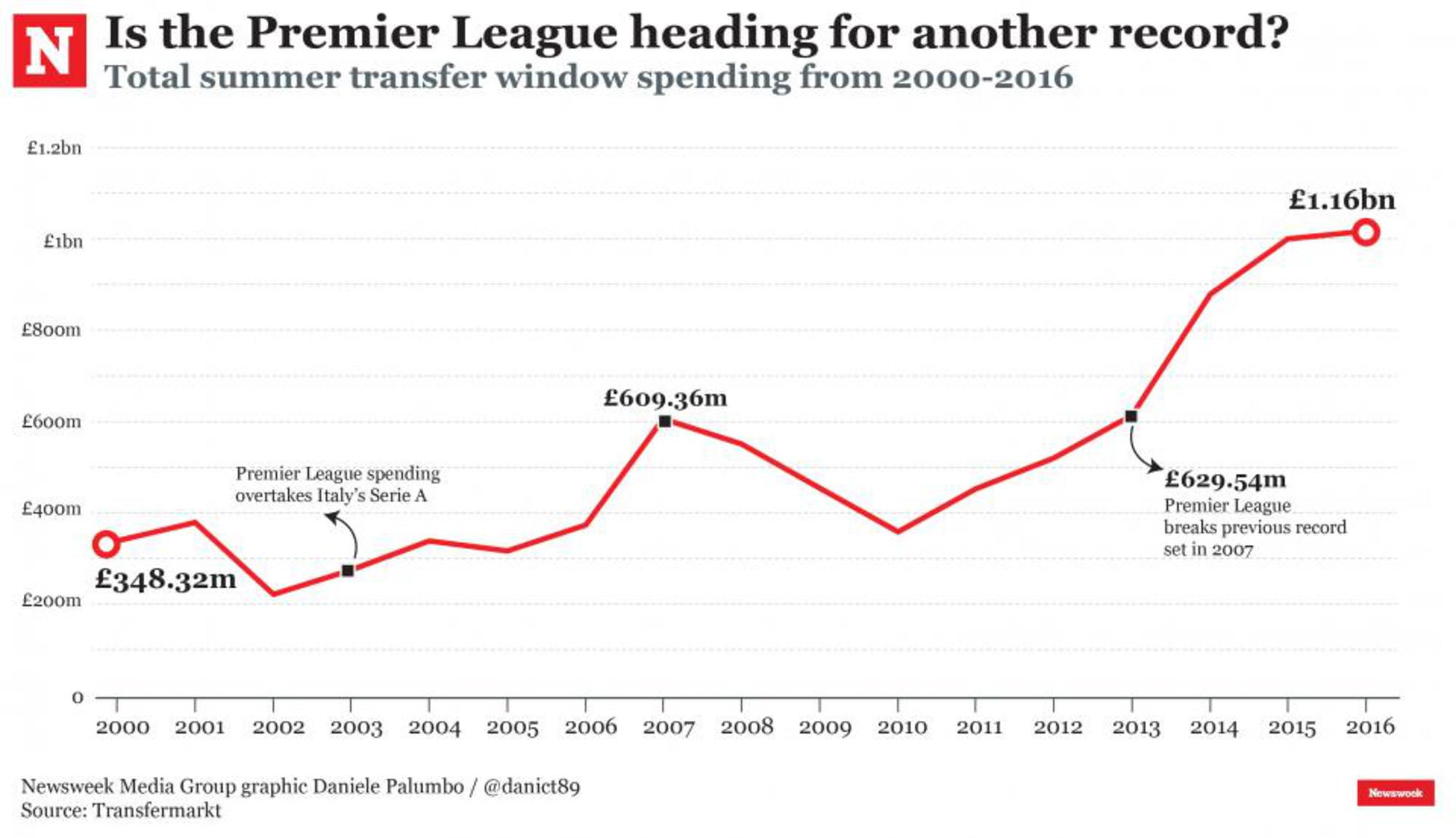}
  \caption{Sample image from the information graphic database used in the filter response experiments.}

\label{fig:sampleimage}
\end{figure}

 For our experiments, we present both quantitative and qualitative results on the interpretability of the sparse coding dictionary versus the CNN kernel filters.    For many of these experiments, we use a sample image from our database, see Figure \ref{fig:sampleimage}, to measure responses.
 
 \subsection{Response Comparison of Similar Filters}

\begin{figure}[t]
  \centering
    \subfigure{\includegraphics[width=2.5cm]{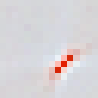}}
    \subfigure{\includegraphics[width=2.5cm]{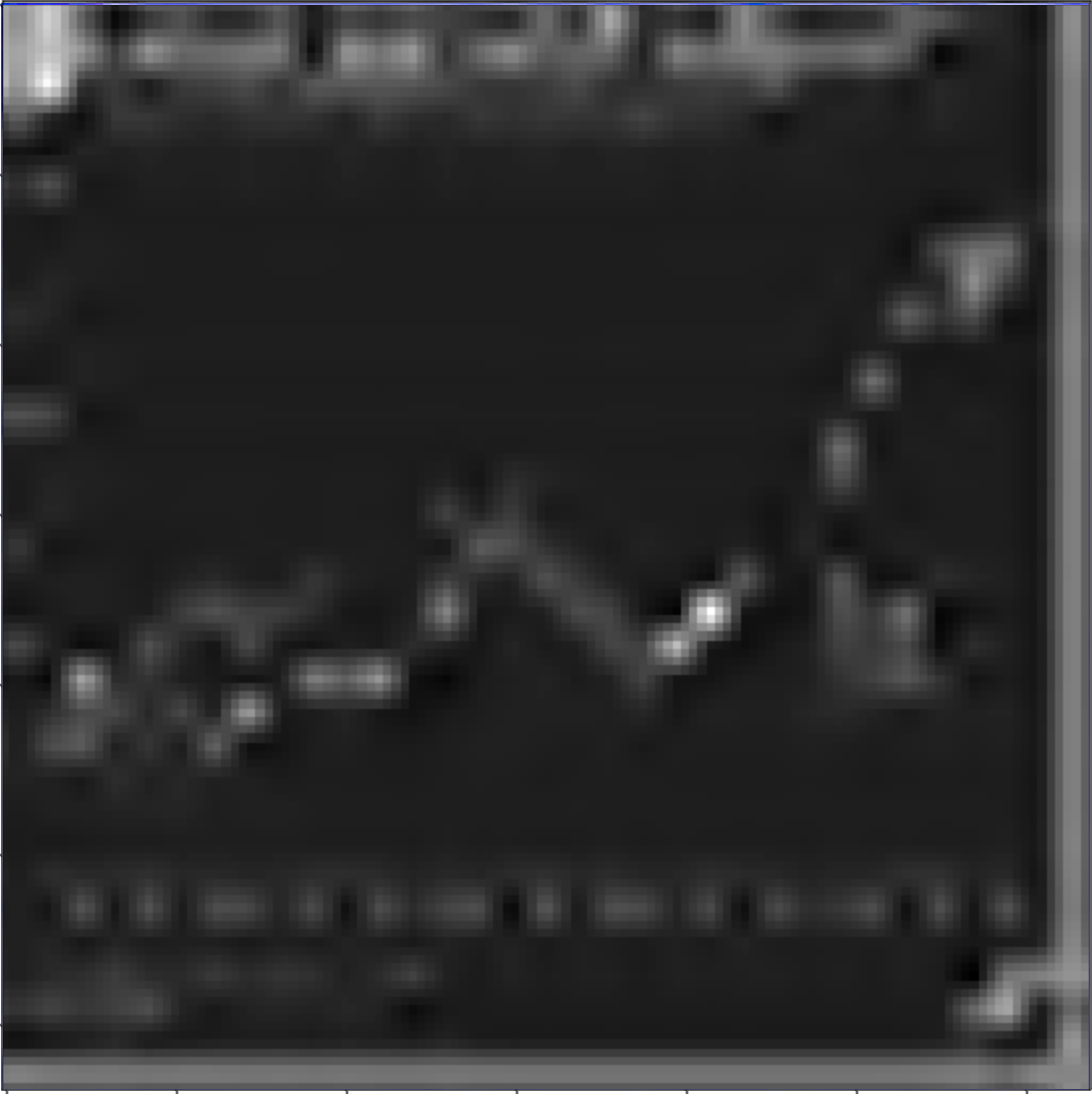}}
    \subfigure{\includegraphics[width=2.5cm]{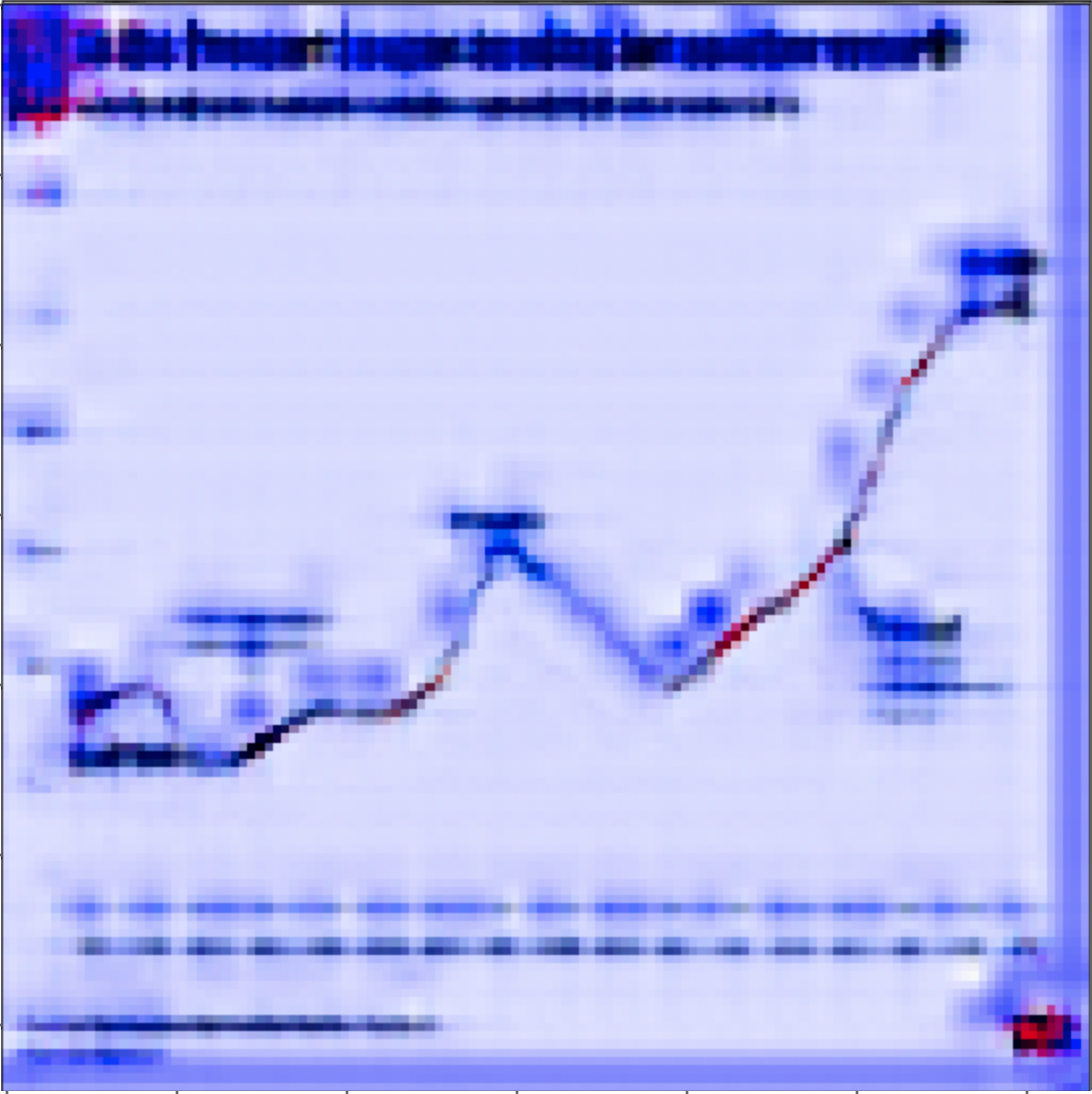}} 
    \setcounter{subfigure}{0}\subfigure[CNN filters]{\includegraphics[width=2.5cm]{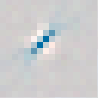}}
    \subfigure[Activation Map]{\includegraphics[width=2.5cm]{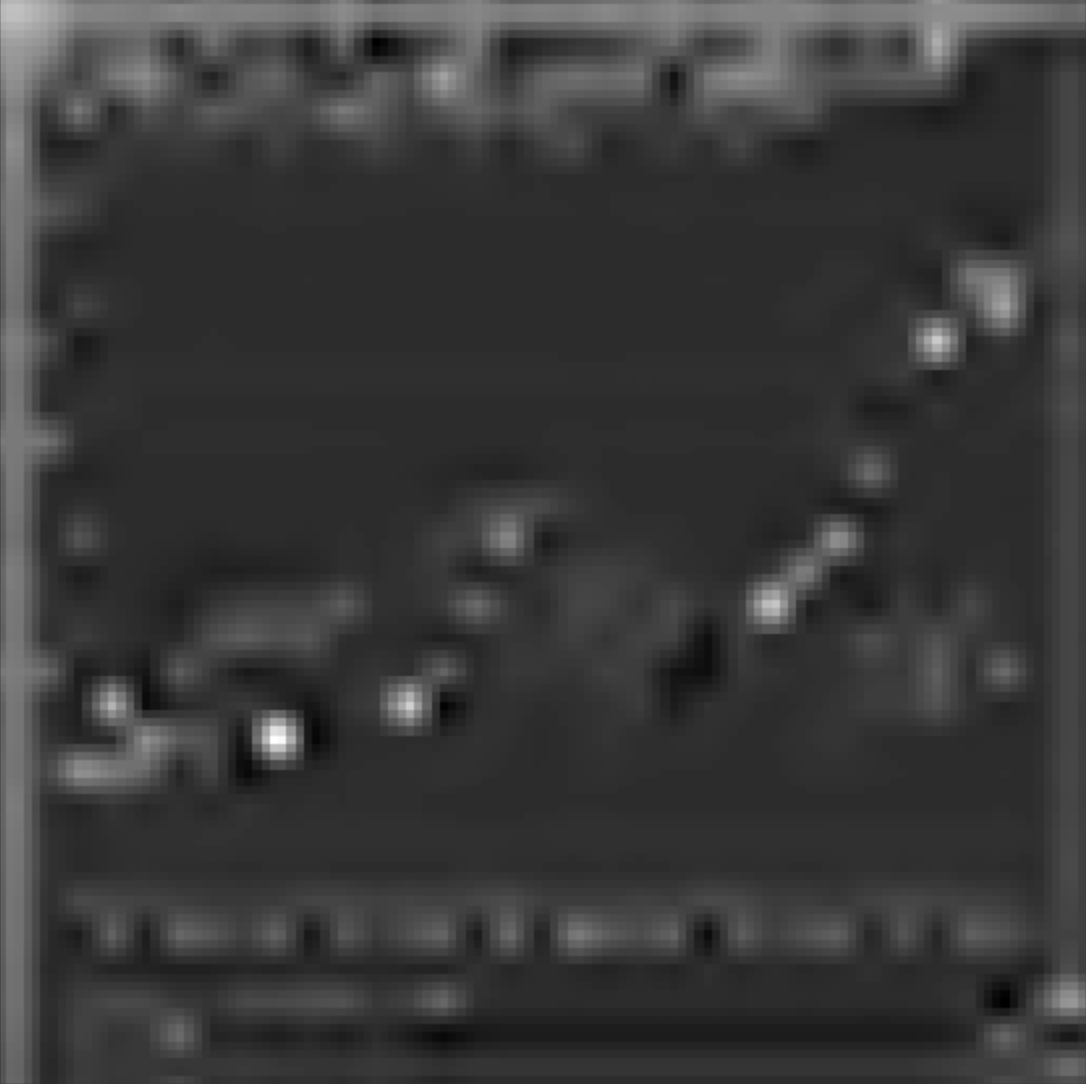}}
    \subfigure[ Overlay]{\includegraphics[width=2.5cm]{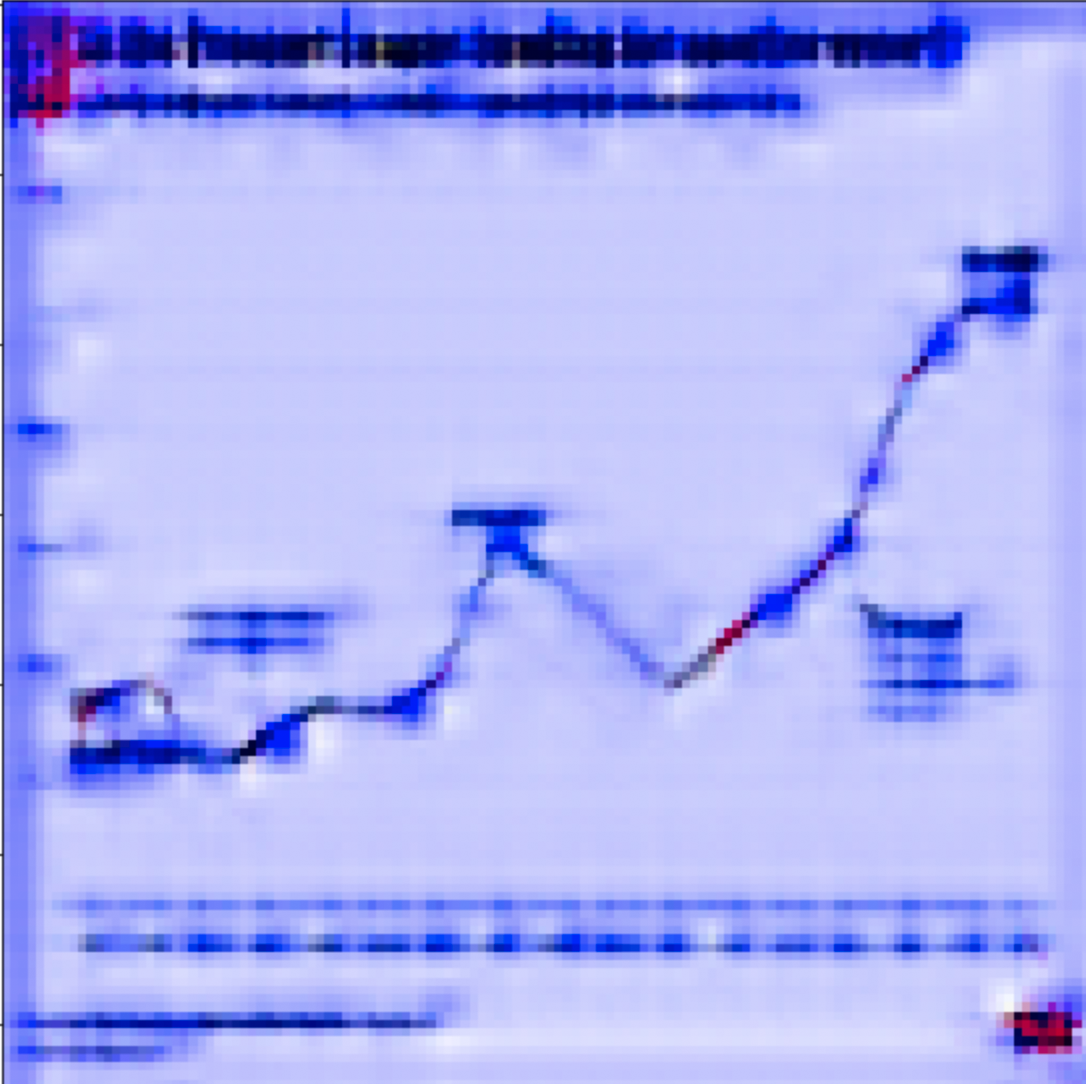}}

  \caption{(a) Selected CNN filters that look like they capture high level information.  (b) The activation maps corresponding to convolving the filter across the image.  (c) An overlay of the activations on the input stimulus.  Note that there are no non-zeros and areas of high frequency seem to activate the filters similarly.}

\label{fig:responses}
\end{figure}

\begin{figure}[t]
    \centering
    \subfigure{\includegraphics[width=2.5cm]{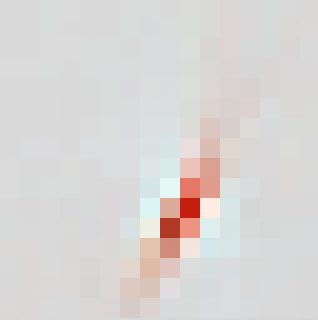}}
    \subfigure{\includegraphics[width=2.5cm]{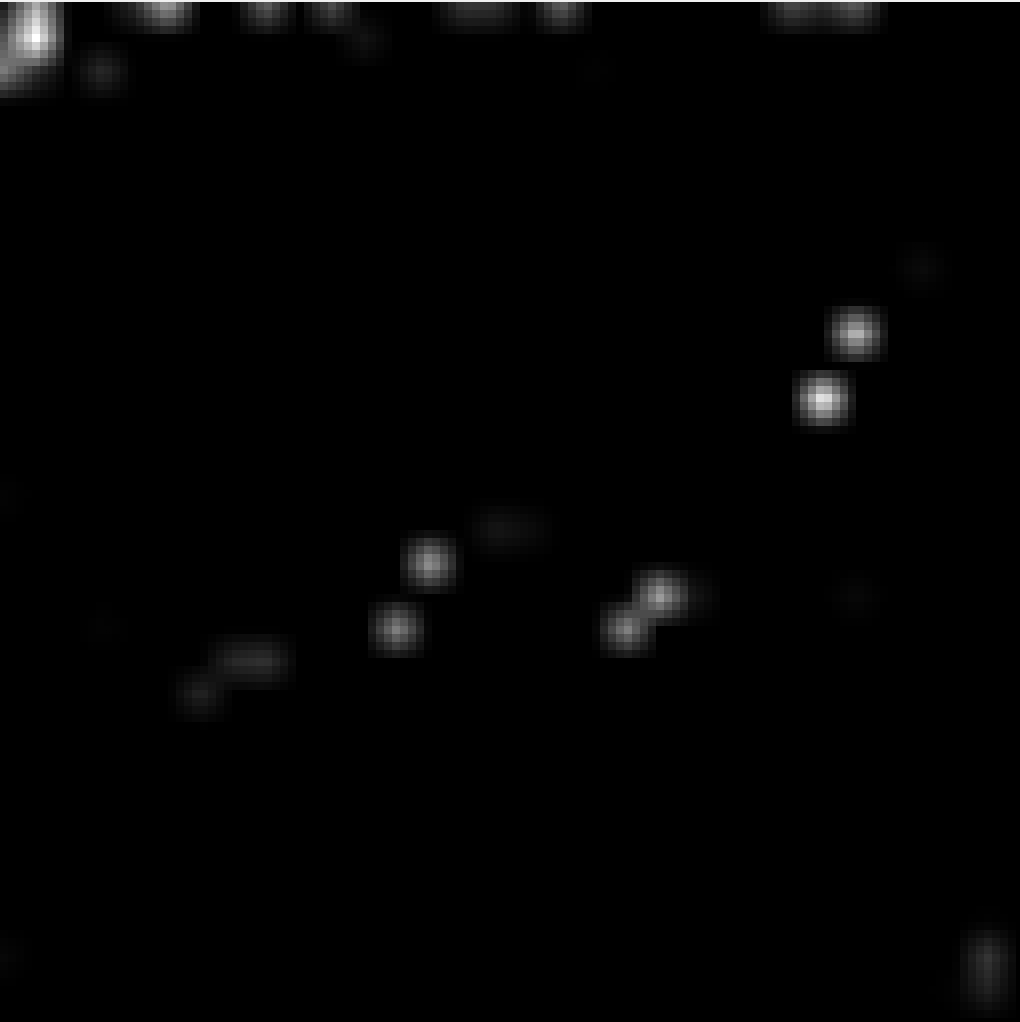}}
    \subfigure{\includegraphics[width=2.5cm]{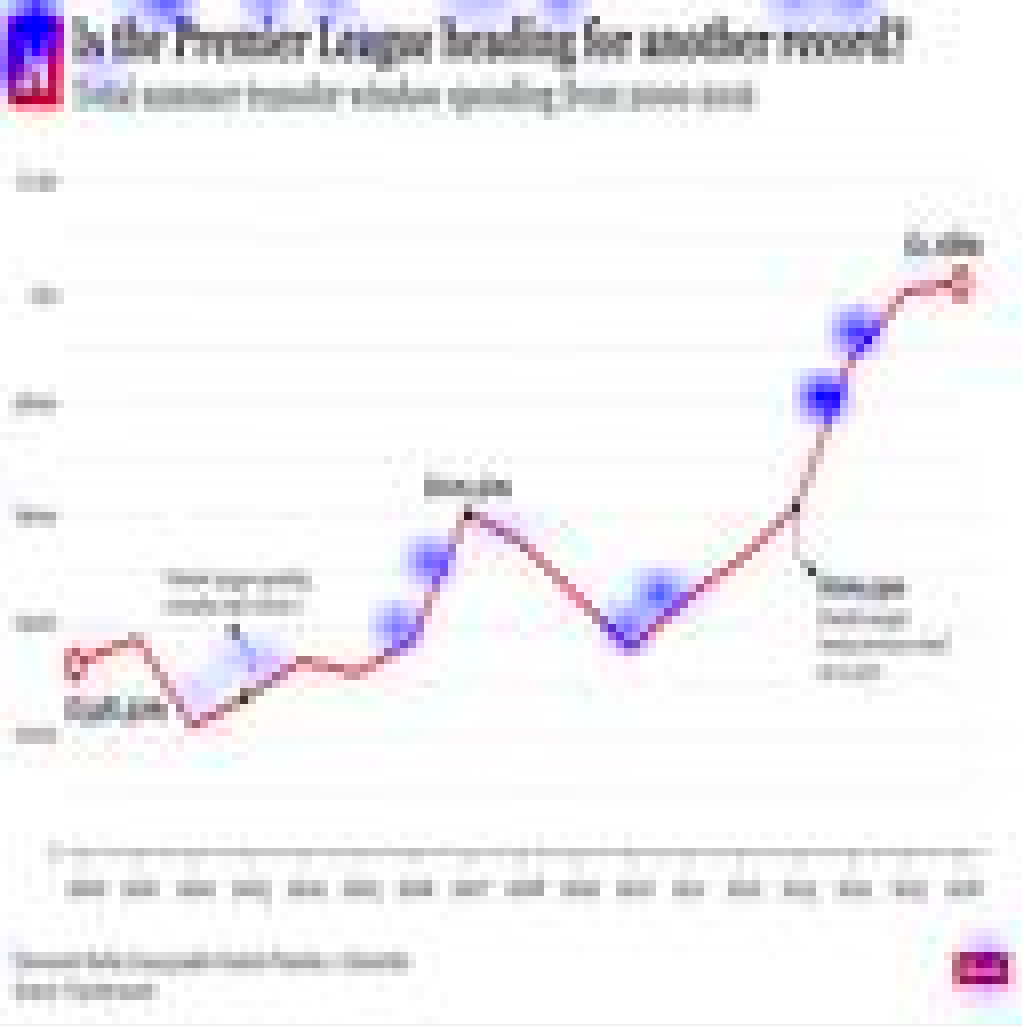}}
    \setcounter{subfigure}{0}\subfigure[Dictionary]{\includegraphics[width=2.5cm]{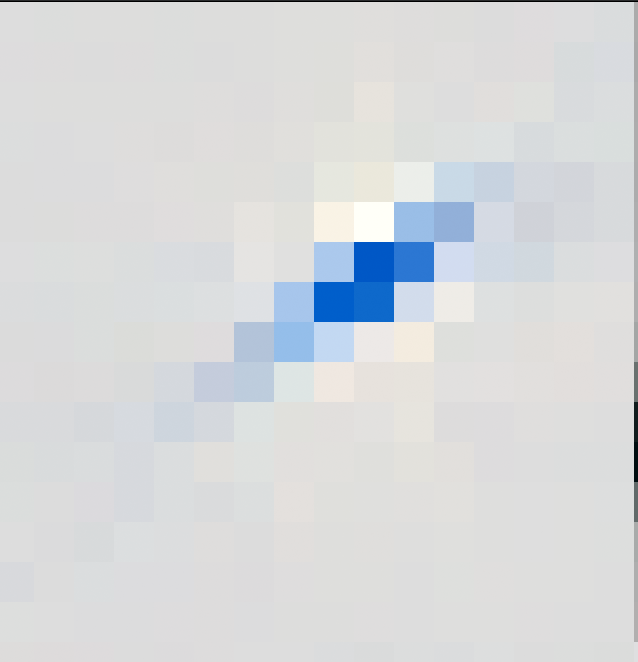}}
    \subfigure[Activation Map]{\includegraphics[width=2.5cm]{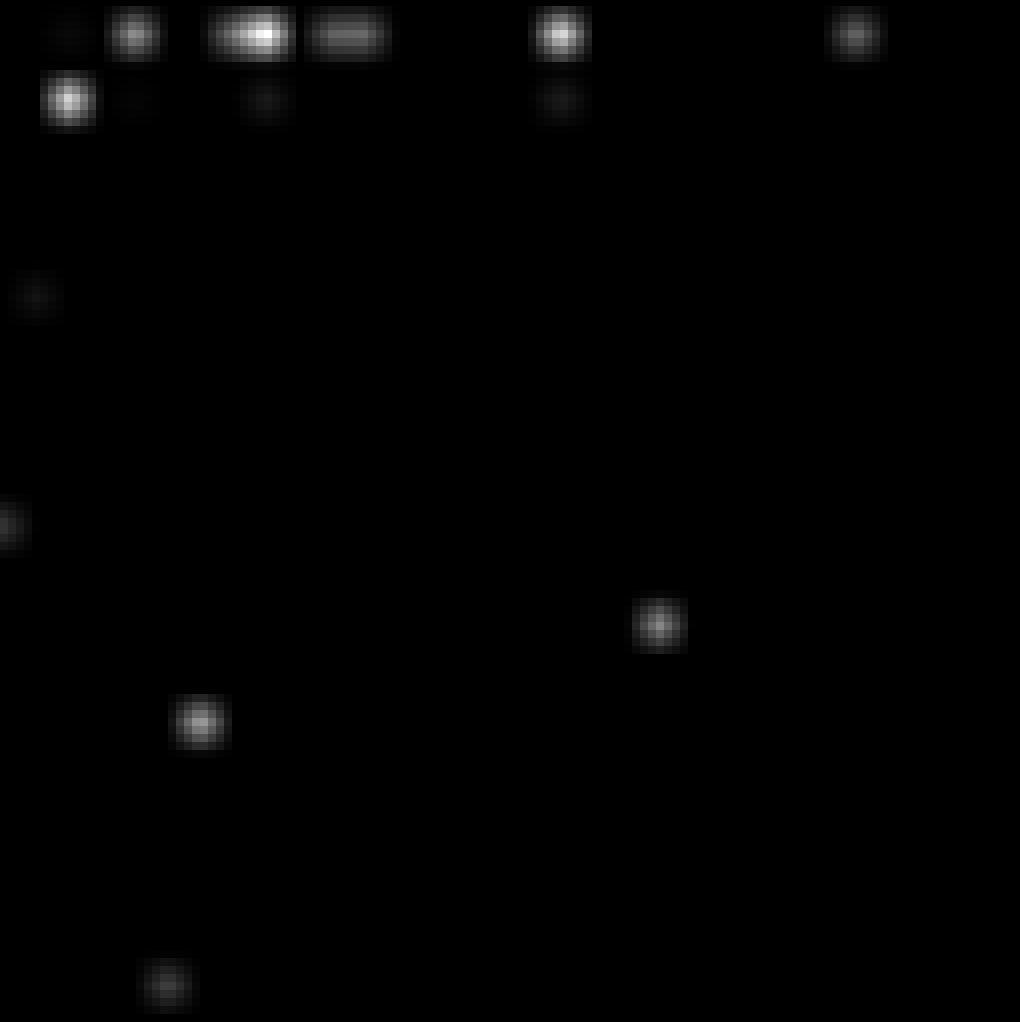}}
    \subfigure[Overlay]{\includegraphics[width=2.5cm]{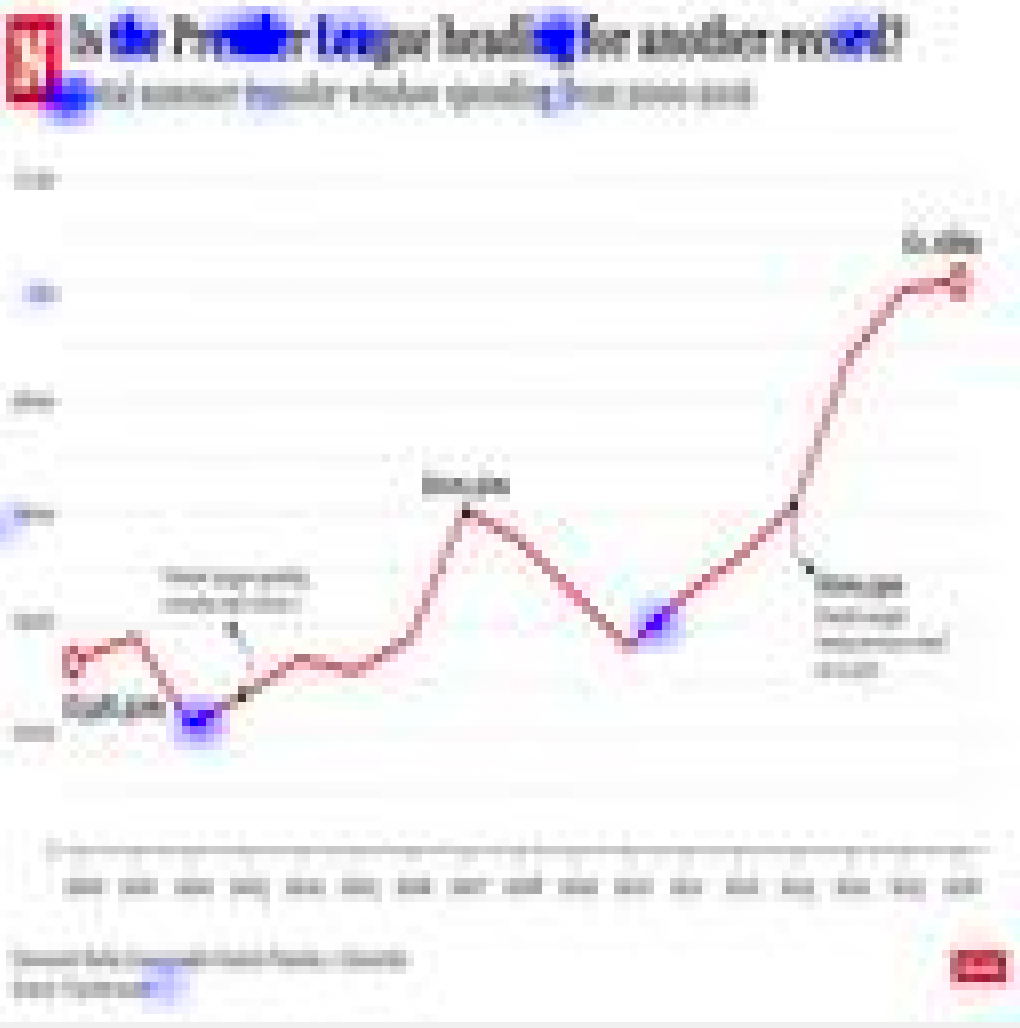}}

  \caption{(a) Selected dictionary elements that look like they capture high level information.  (b) The activation maps corresponding to sparse coding the image.  (c) An overlay of the activations on the input stimulus.  Note that there are many zero responses and the dictionary elements are selectively active where the image patch and dictionary element have high correlation.}

\label{fig:sparseresponses}
\end{figure}

 We then visually inspect the dictionary elements in our sparse coding framework, and the image filters learned from the CNN, and find features that are visually similar.  In this case, we selected a red diagonal line and diagonal blue line shown in Figure \ref{fig:responses}(a) and Figure \ref{fig:sparseresponses}(a).  Here, we are able to see the contrast of the output of the different methods on the corresponding activation maps.  Upon initial inspection, the CNN seems to respond strongly in the areas within the image that have high correlation with the filter; however, after a closer evaluation, it is apparent that the CNN responds somewhat indiscriminately to areas of high frequency, and only slightly more so in image areas that have high correlation to the image filter.  In fact, the response of a red line segment and blue line segment CNN filter seem to have nearly identical responses.
 
 In contrast, the response of the corresponding sparse coding dictionary elements is extremely sparse and selective.  Furthermore, the areas within the image of high correlation to the dictionary element selectively activates the element, and is otherwise silent with zero response.  This behavior of the CNN filters and dictionary elements can been seen in Figure \ref{fig:responses} and Figure \ref{fig:sparseresponses}.

\subsection{Response Comparison of an Image Patch}
\begin{figure*}[th]
  \centering
    \subfigure[Image Patch]{\includegraphics[width=3.2cm]{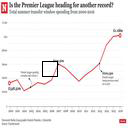}}
    \subfigure[Sparse Dictionary]{\includegraphics[width=3.4cm]{images/V1ToInputErrorW00950000.png}}
    \subfigure[Sparse Activation]{\includegraphics[width=3.4cm]{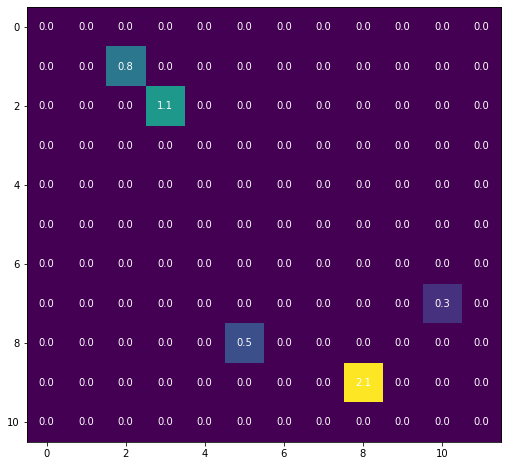}}
    \subfigure[CNN Filters]{\includegraphics[width=3.4cm]{images/ae175_05noise.png}}
    \subfigure[Dense Activation]{\includegraphics[width=3.4cm]{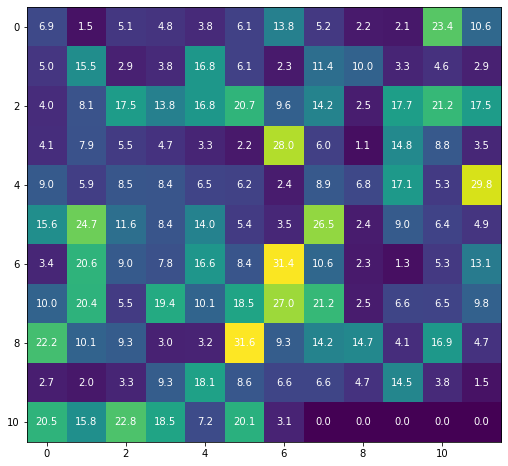}}
  \caption{(a) A 16x16 image patch.  (b) The converged sparse coding dictionary.  (c) The coefficients of the activated neurons in the sparse coding dictionary.  (d) The CNN filters. (e) The dense activation from convolving the CNN filters.}

\label{fig:patch2}
\end{figure*}

In this next set of experiments, we examine a specific image patch and visualize the response of each of the 128 dictionary elements.  We take an area of interest shown in Figure \ref{fig:patch2}(a) and compute the coefficients of the CNN filters and dictionary elements needed to faithfully reconstruct that image patch.  In the CNN autoencoder case, there are no non-zero elements; every image filter has a response, see Figure \ref{fig:patch2}(e).  However, in the sparse coding case, there are only 5 out of 128 non-zero elements.  The coefficients can be seen in Figure \ref{fig:patch2}(c).

  \begin{figure}[ht]
    \subfigure[]{\includegraphics[width=2cm]{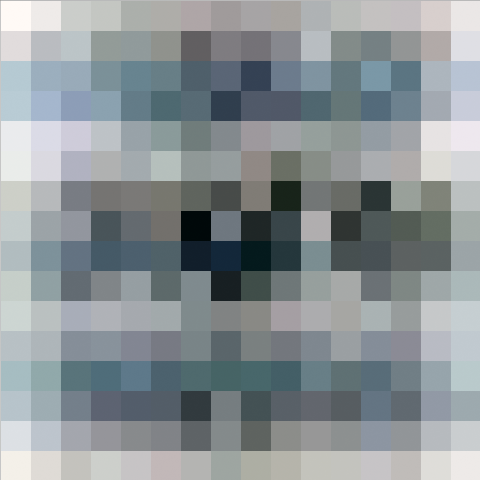}}
    \subfigure[]{\includegraphics[width=2cm]{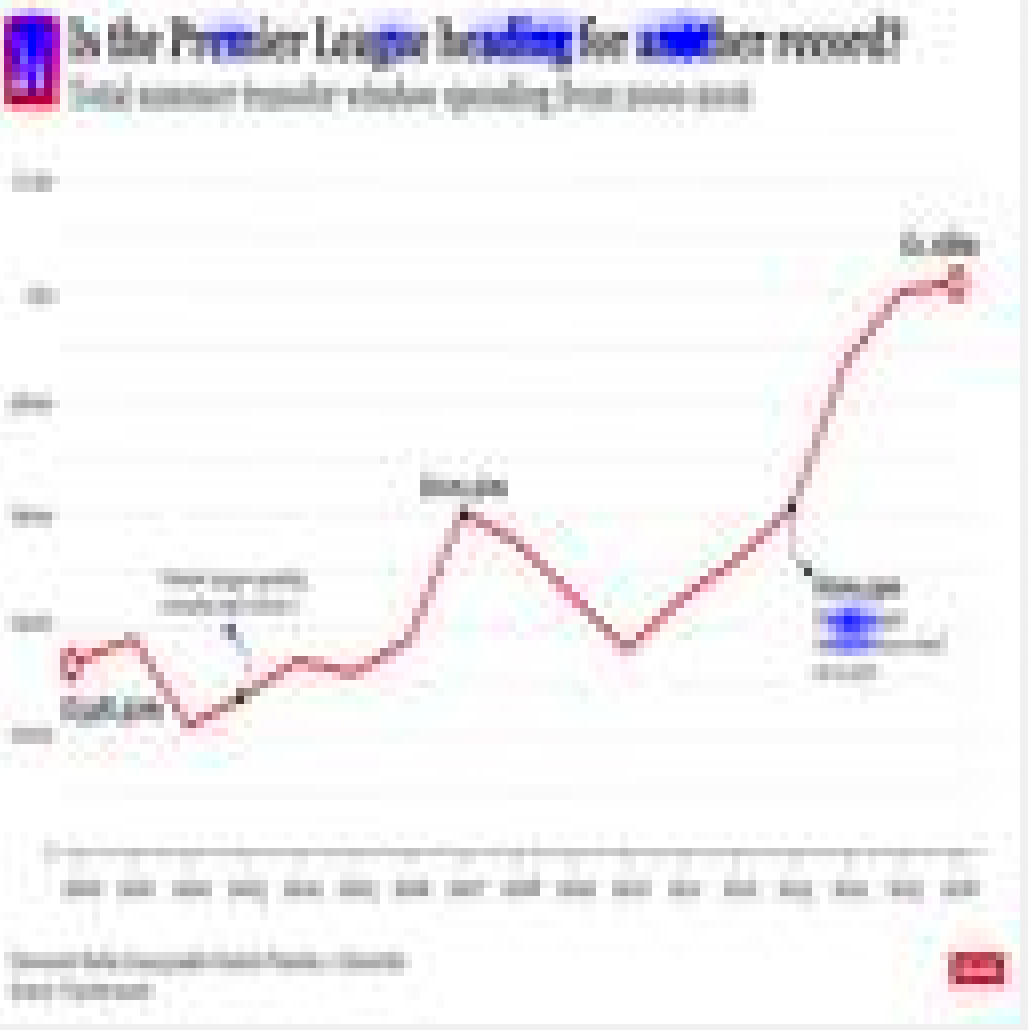}}
    \subfigure[]{\includegraphics[width=2cm]{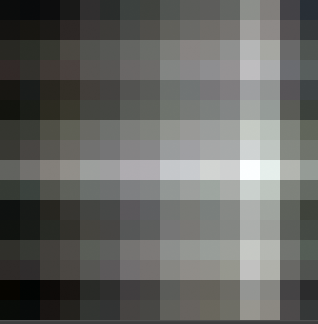}}
    \subfigure[]{\includegraphics[width=2cm]{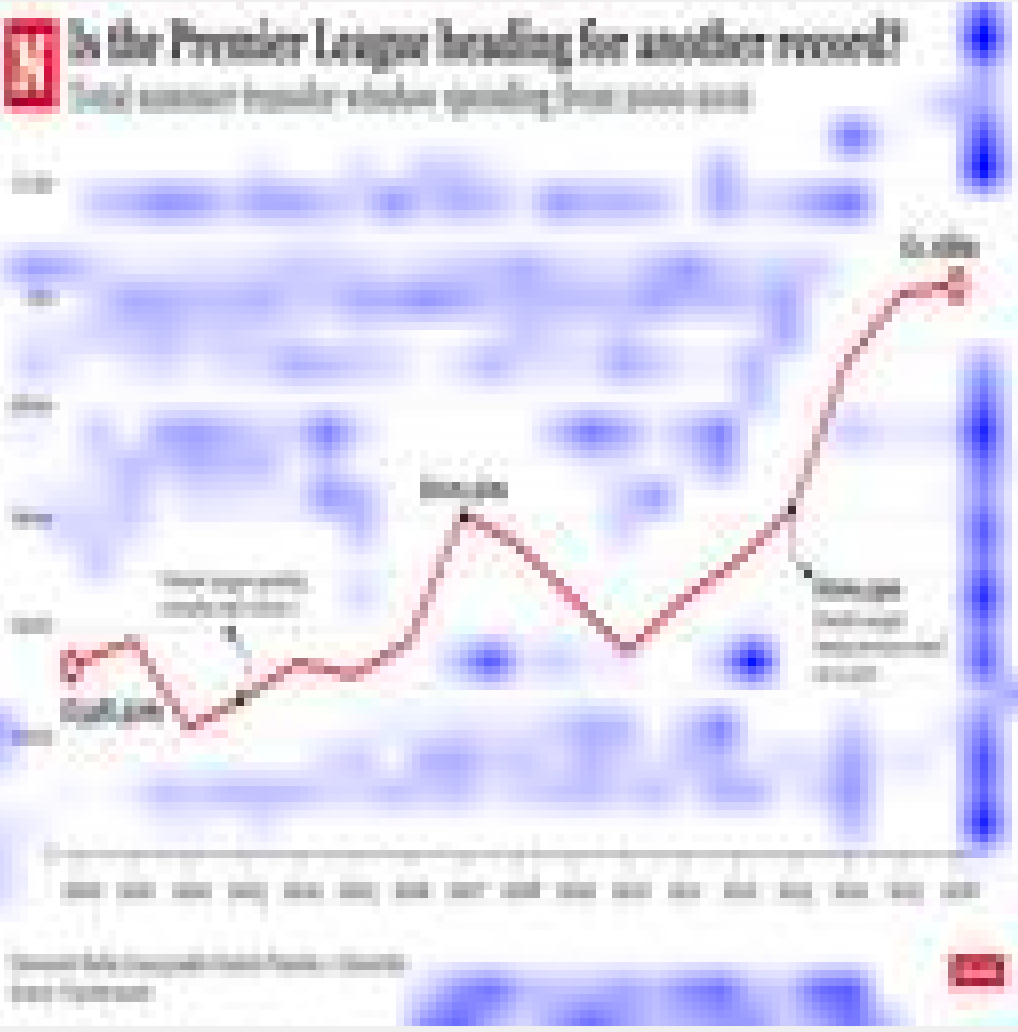}}

  \caption{Dictionary elements (a) and (c) and the respective areas that they respond to in the image (b) and (d).  (a) Responds to multi-line text or large font text, and (c) responds to empty space. }

\label{fig:moreex}
\end{figure}

  \begin{figure}[ht]
    \subfigure[Percent Active]{\includegraphics[width=4cm]{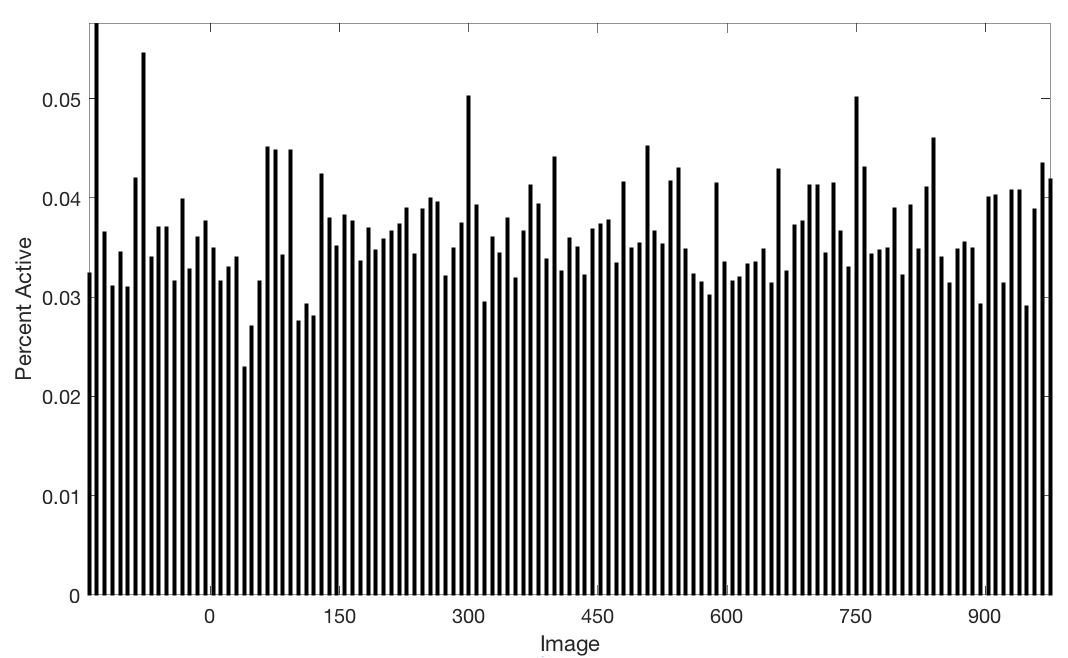}}
        \subfigure[Frequency of Usage]{\includegraphics[width=4cm]{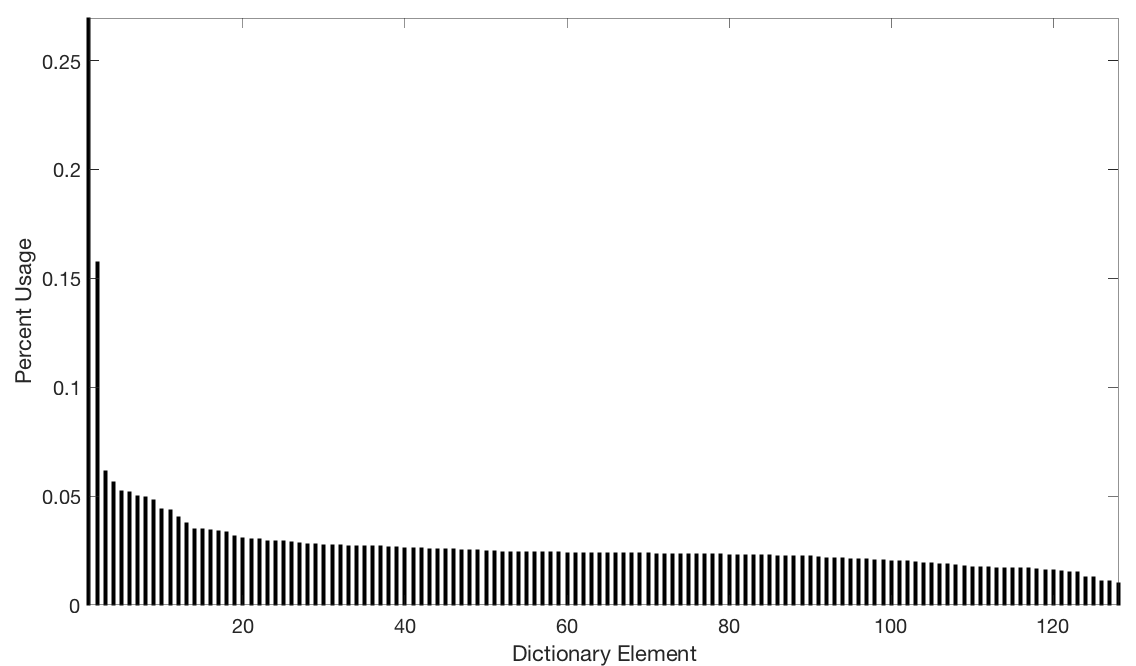}}
  \caption{Percent active and frequency of usage of the sparse coding dictionary.  Generally the activity is less than 4\% and the frequency of usage is less than 5\%.}

\label{fig:bargraph}
\end{figure}

\subsection{Intra and Inter Cross Correlation}
In our last set of experiments, we analyze the response across the entire dataset of 1,000 information graphics.  In order to determine the sparsity and selectivity of the dictionary and convolutional filters, we compute the cross correlation of the activated responses both intraimage (the cross correlation of the activations from different filters on the same image) and interimage (the cross correlation of the activations across different images).   We make sure that we normalize the weights such that they have comparable responses magnitudes.  We present the first and second order moments in Table \ref{tab:results}.  Our sparse coding method has more than an order of magnitutde less cross correlation across the dataset, and a smaller intraimage standard deviation.  Conceptually, this can be interpreted as most of the activation maps are decorrelated and hold distinct information; whereas, high correlation in the CNN case reflects non-specific responses and entagled contributions burried within the output activity.

\begin{center}
\begin{table}[tbh]
    \footnotesize
    \begin{tabular}{ p{2.75cm}  l  l  l  }
    \hline
    \textbf{Method}  & \textbf{Mean} & \textbf{Intra-std} & \textbf{Inter-std} \\ \hline 
    CNN Filter &  420.80 & 259.63 & 15.06\\ 
    Sparse Coding & 29.54 & 118.35 & 89.44\ \\
  
    \end{tabular}
    	\caption{Mean and standard deviation within image and across the dataset for the CNN activations and sparse coding activations.  Sparse coding has an order of magnitude less average cross correlation on our dataset. }
	\label{tab:results}
	\end{table}
\end{center}

\subsection{Sparse Coding Dictionary} 
In addition to being highly selective, our sparse coding dictionary creates more interpretable representations than the CNN counterpart.  In Figure \ref{fig:moreex}, we illustrate several other cases where we can directly correlate the dictionary element with a semantic category.  Furthermore, in Figure \ref{fig:bargraph}, we can see the distribution of active elements across the entire dataset, as well as the frequency of usage of a particular dictionary element.  The percent active is generally around 4\% for an image, and most dictionary elements are generally used less than 5\% of the time.
\section{Conclusion}

Oftentimes, deep-learning techniques are thought of as black boxes, or highly uninterpretable models.  Instead of building parallel explainable models or modules on top of deep learning, we discuss a biologically plausible neural implementation.  We propose that elements of sparsity, competition, and selectivity are important for interpretability.  Each portion of the data is accounted for by a sparse set of neurons, and thus the contribution of any given neuron to the explanation of the input data can be directly determined.  Such explanations in deep learning models are not possible because the same portion of the data can be represented by many neurons and thus the explanation for any given neuron's activity becomes highly confounded.
 \bibliographystyle{aaai}
 \bibliography{egbib}

\begin{thebibliography}{}

\bibitem[\protect\citeauthoryear{Field}{1994}]{field1994goal}
Field, D.~J.
\newblock 1994.
\newblock What is the goal of sensory coding?
\newblock {\em Neural computation} 6(4):559--601.

\bibitem[\protect\citeauthoryear{Fukushima}{1980}]{fukushima1980}
Fukushima, K.
\newblock 1980.
\newblock Neocognitron: A self-organizing neural network model for a mechanism
  of pattern recognition unaffected by shift in position.
\newblock {\em Biological Cybernetics} 36(4):193--202.

\bibitem[\protect\citeauthoryear{Hopfield}{1984}]{hopfield1984neurons}
Hopfield, J.~J.
\newblock 1984.
\newblock Neurons with graded response have collective computational properties
  like those of two-state neurons.
\newblock {\em Proceedings of the national academy of sciences}
  81(10):3088--3092.

\bibitem[\protect\citeauthoryear{Hubel and Wiesel}{1962}]{hubel1962receptive}
Hubel, D.~H., and Wiesel, T.~N.
\newblock 1962.
\newblock Receptive fields, binocular interaction and functional architecture
  in the cat's visual cortex.
\newblock {\em The Journal of physiology} 160(1):106--154.

\bibitem[\protect\citeauthoryear{Itti and Koch}{2001}]{itti2001feature}
Itti, L., and Koch, C.
\newblock 2001.
\newblock Feature combination strategies for saliency-based visual attention
  systems.
\newblock {\em Journal of Electronic imaging} 10(1):161--169.

\bibitem[\protect\citeauthoryear{Kim and McCoy}{2018}]{kim2018multimodal}
Kim, E., and McCoy, K.~F.
\newblock 2018.
\newblock Multimodal deep learning using images and text for information
  graphic classification.
\newblock In {\em Proceedings of the 20th International ACM SIGACCESS
  Conference on Computers and Accessibility},  143--148.

\bibitem[\protect\citeauthoryear{Kim, Onweller, and
  McCoy}{2020}]{kim2020multimodal}
Kim, E.; Onweller, C.; and McCoy, K.~F.
\newblock 2020.
\newblock Information graphic summarization using a collection of multimodal
  deepneural networks.
\newblock In {\em International Conference on Pattern Recognition, ICPR}.

\bibitem[\protect\citeauthoryear{Knudsen}{2007}]{knudsen2007fundamental}
Knudsen, E.~I.
\newblock 2007.
\newblock Fundamental components of attention.
\newblock {\em Annu. Rev. Neurosci.} 30:57--78.

\bibitem[\protect\citeauthoryear{Le \bgroup et al\mbox.\egroup
  }{2012}]{quocle2012}
Le, Q.~V.; Ranzato, M.; Monga, R.; Devin, M.; Chen, K.; Corrado, G.~S.; Dean,
  J.; and Ng, A.~Y.
\newblock 2012.
\newblock Building high-level features using large scale unsupervised learning.

\bibitem[\protect\citeauthoryear{Lecun, Bengio, and Hinton}{2015}]{lecun2015}
Lecun, Y.; Bengio, Y.; and Hinton, G.
\newblock 2015.
\newblock Deep learning.
\newblock {\em Nature Cell Biology} 521(7553):436--444.

\bibitem[\protect\citeauthoryear{LeCun \bgroup et al\mbox.\egroup
  }{1989}]{lecun1989}
LeCun, Y.; Boser, B.; Denker, J.~S.; Henderson, D.; Howard, R.~E.; Hubbard, W.;
  and Jackel, L.~D.
\newblock 1989.
\newblock Backpropagation applied to handwritten zip code recognition.
\newblock {\em Neural Computation} 1(4):541--551.

\bibitem[\protect\citeauthoryear{Miller}{2019}]{miller2019explanation}
Miller, T.
\newblock 2019.
\newblock Explanation in artificial intelligence: Insights from the social
  sciences.
\newblock {\em Artificial Intelligence} 267:1--38.

\bibitem[\protect\citeauthoryear{Olshausen and
  Field}{1997}]{olshausen1997sparse}
Olshausen, B.~A., and Field, D.~J.
\newblock 1997.
\newblock Sparse coding with an overcomplete basis set: A strategy employed by
  v1?
\newblock {\em Vision research} 37(23):3311--3325.

\bibitem[\protect\citeauthoryear{Rozell \bgroup et al\mbox.\egroup
  }{2007}]{rozell2007locally}
Rozell, C.; Johnson, D.; Baraniuk, R.; and Olshausen, B.
\newblock 2007.
\newblock Locally competitive algorithms for sparse approximation.
\newblock In {\em Image Processing, 2007. ICIP 2007. IEEE International
  Conference on}, volume~4,  IV--169.
\newblock IEEE.

\bibitem[\protect\citeauthoryear{Rudin}{2019}]{rudin2019stop}
Rudin, C.
\newblock 2019.
\newblock Stop explaining black box machine learning models for high stakes
  decisions and use interpretable models instead.
\newblock {\em Nature Machine Intelligence} 1(5):206--215.

\bibitem[\protect\citeauthoryear{Shoham, O’Connor, and
  Segev}{2006}]{shoham2006silent}
Shoham, S.; O’Connor, D.~H.; and Segev, R.
\newblock 2006.
\newblock How silent is the brain: is there a “dark matter” problem in
  neuroscience?
\newblock {\em Journal of Comparative Physiology A} 192(8):777--784.

\bibitem[\protect\citeauthoryear{Vincent \bgroup et al\mbox.\egroup
  }{2010}]{vincent2010stacked}
Vincent, P.; Larochelle, H.; Lajoie, I.; Bengio, Y.; Manzagol, P.-A.; and
  Bottou, L.
\newblock 2010.
\newblock Stacked denoising autoencoders: Learning useful representations in a
  deep network with a local denoising criterion.
\newblock {\em Journal of machine learning research} 11(12).

\end{thebibliography}
\end{document}